\documentclass[twocolumn,10pt]{asme2ej}

\usepackage{epsfig} 

\usepackage{subcaption}
\usepackage{graphics}
\usepackage{bm}
\usepackage{amssymb}
\usepackage{xcolor}
\usepackage{amsmath}
\usepackage{url}
\usepackage{textcomp}
\newcommand{\comment}[1]{}

\newtheorem{definition}{Definition} 

\title{Dynamic Simulation-Guided Design of Tumbling Magnetic Microrobots}

\author{ \parbox{3 in}{\centering Jiayin Xie\\
         Department of Mechanical Engineering\\
         Stony Brook University\\
         Stony Brook, NY 11794\\
         {\tt\small jiayin.xie@stonybrook.edu}}
         \hspace*{ 0.0 in}
         \parbox{3 in}{\centering Chenghao Bi\\
         School of Mechanical Engineering\\
         Purdue University\\
         West Lafayette, IN 47907\\
         {\tt\small bi10@purdue.edu}}\\
}

\author{ \parbox{3 in}{\centering David J. Cappelleri\\
         School of Mechanical Engineering\\
         Purdue University\\
         West Lafayette, IN 47907\\
         {\tt\small dcappell@purdue.edu}}
         \hspace*{ 0.0 in}
         \parbox{3 in}{\centering Nilanjan Chakraborty \thanks{Address all correspondence to this author.}\\
         Department of Mechanical Engineering\\
         Stony Brook University\\
         Stony Brook, NY 11794\\
         {\tt\small nilanjan.chakraborty@stonybrook.edu}}
} 

\begin{document}

 \maketitle

\begin{abstract}
{\it Design of robots at the small scale is a trial-and-error based process, which is costly and time-consuming. There are few dynamic simulation tools available to accurately predict the motion or performance of untethered microrobots as they move over a substrate. At smaller length scales, the influence of adhesion and friction, which scales with surface area, becomes more pronounced. Thus, rigid body dynamic simulators, which implicitly assume that contact between two bodies can be modeled as point contact are not suitable. In this paper, we present techniques for simulating the motion of microrobots where there can be intermittent and non-point contact between the robot and the substrate. We use these techniques to study the motion of tumbling microrobots of different shapes and select shapes that are optimal for improving locomotion performance. Simulation results are verified using experimental data on linear velocity, maximum climbable incline angle, and microrobot trajectory. Microrobots with improved geometry were fabricated, but limitations in the fabrication process resulted in unexpected manufacturing errors and material/size scale adjustments. The developed simulation model is able to incorporate these limitations and emulate their effect on the microrobot's motion, reproducing the experimental behavior of the tumbling microrobots, further showcasing the effectiveness of having such a dynamic model. }
\end{abstract}





\section{Introduction}
\begin{figure}
\centering
\includegraphics[width=2in]{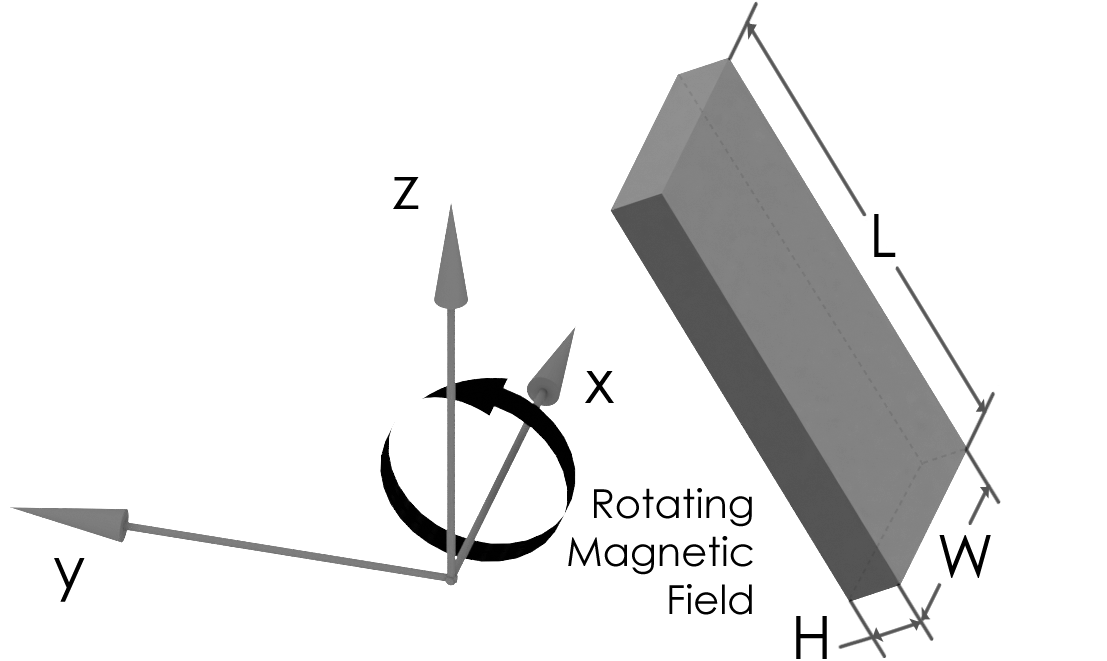}
\caption{Microscale magnetic tumbling ($\mu$TUM) robot tumbles on a planar surface. A magnetic field rotating counterclockwise about the $x$ axis  causes the robot to rotate about the same direction and tumble forward (along the direction of $y$ axis). The length, width, and height of the $\mu$TUM robot are L, W, and H respectively. } 
\label{figure_robot} 
\end{figure}

Tumbling microrobots have the potential to go to previously unreachable areas of the body and perform tasks such as targeted drug delivery, tissue biopsies, and toxin neutralization~\cite{ErkocYC+18}. Magnetically actuated microrobots that use the difference in the orientation of the robot$'$s internal magnetization and that of a rotating external magnetic field to induce a torque on the robot, and make it tumble forward end-over-end, have been proposed in the literature~\cite{Jing2013b, Jing2013c, Bi2018}. Figure~\ref{figure_robot} shows the schematic sketch of such a tumbling microrobot. The tumbling locomotion has been shown to be versatile in both wet and dry environments, on steep inclines, and on rough surfaces~\cite{Bi2018}. The external magnetic fields actuating the robot harmlessly penetrate living tissue and allow for tetherless locomotion.  One key limitation of external magnetic fields, however, is that they decrease volumetrically in strength as distance increases between the magnetic target and the source of the field. Therefore, it is beneficial to optimize the robot's design to achieve the most mobility under limited magnetic field strengths. It is also beneficial to optimize the design to travel over as many different surfaces as possible. 

Currently, the design of tumbling microrobots is a trial-and-error based process, which is costly and time consuming. Thus, a flexible dynamic simulation tool for virtual design iteration and optimization would be highly valuable. Therefore, {\em our goal in this paper is to create a dynamic simulation tool that can be used to study the motion of microrobots of different geometry and manufacture a subset of the robots for experimentation}. This will greatly help reduce the cost and effort of the microrobot design process, as the designer can hone in on the most promising designs.


A critical challenge for simulating the tumbling microrobot is to model the intermittent and non-point contact between the robot and substrate, which will change during the motion based on the contact mode. For example, for the box-like microrobot (shown in Figure~\ref{figure_robot}), tumbling over a flat surface will alternate between area contact and line contact as it flips end over end. Furthermore, the contact area will change depending on the face that is in contact. This is crucial because, at the length scale of microrobots, the dynamics is heavily influenced by adhesive forces that scale with the surface area.
Most existing dynamic simulation methods for microrobots~\cite{Pawashe2008,Pawashe2009a} implicitly assume that the contact between two bodies can be modeled as point contact. They choose contact points {\em a priori} in an ad hoc manner to represent the contact patch. For the tumbling robot, since the contact patch is time-varying it is not possible to choose contact point {a priori} and would thus introduce inaccuracy in simulation. Recently, we developed principled methods~\cite{xie2016rigid,Xie2018,xie2019}, to simulate intermittently contacting rigid bodies with planar convex and non-convex contact patches. In this paper, based on our previous work, we develop a method for simulating the motion of microrobots where the contact between the robot and the substrate is intermittent and possibly non-point. 

{\bf Contributions}: Our contributions are as follows: (a) We extend our dynamic model in\cite{xie2016rigid} to handle the magnetic torque induced from a rotational magnetic field and the surface area-dependent adhesive forces acting on a rigid body microrobot. (b) We describe a procedure to compute the adhesive forces, which will change during motion based on the contact mode. (c) We present numerical simulation results and perform comparisons with experiments in several scenarios. (d) We simulate the motion of microrobots with more complex, untested geometry and find the shapes that best improve locomotion performance in terms of linear velocity and maximum climbable incline angle. (e) We build upon earlier work described in~\cite{XieB+2019} by fabricating the optimal microrobot geometries and experimentally characterizing the effects of manufacturing errors on the resulting motion behavior.  (f) Additionally, we demonstrate the capabilities of the developed model by incorporating the identified manufacturing errors into it and successfully reproduce the altered motion behavior within the simulation.

A preliminary version of this paper appeared in~\cite{XieB+2019}. In this paper, we have rewritten some sections of ~\cite{XieB+2019} for clarity. Furthermore, we have added an extensive simulation analysis of the design space and reported additional experimental studies supporting the simulation studies.



{\bf Outline of the paper}: This paper starts by discussing related work on dynamic models in Section 2 with additional comments on microscale effects. Section 3 then provides a high level overview of the design space explored in this work. In Section 4, our general dynamic model is described in detail and further refined for tumbling microrobots in Section 5. Next, in Section 6, we verify the results of the dynamic model with a variety of experimental locomotion tests. Performance predictions are made in Section 7 for four alternative tumbling microrobot geometries and Section 8 describes the manufacturing process and limitations for fabricating the two best performing designs. Section 9 discusses the incorporation of manufacturing errors into the dynamic model and compares the altered motion prediction results with experimental data from the two alternative microrobot designs. Conclusions about tumbling microrobot performance and the capabilities of the dynamic model are discussed in Section 10. Finally, concluding thoughts and a future outlook are described in Section 11.

\vspace{-0.20in}
\section{Related Work}

Past literature has demonstrated dynamic models for several mobile microrobots. Pawashe \textit{et al.} simulated a planar microrobot with stick-slip motion over dry horizontal surfaces \cite{Pawashe2008,Pawashe2009a}. The simulation was able to predict the robot's orientation and linear velocity over time under various external field parameters and surface properties. However, this model does not consider tumbling locomotion with non-point contact patches. Hu \textit{et al.} developed models for predicting the velocities of the rolling, walking, and crawling gaits of a soft-bodied magnetic millibot capable of multimodal locomotion~\cite{Hu2018}. The model helped determined which geometric dimensions were critical for the success of particular gaits of the robot. Morozov \textit{et al.} proposed a general theory to study the dynamics of arbitrarily-shaped magnetic propellers and rationalize previously unexplained experimental observations~\cite{Morozov2017}. To date, a comprehensive three-dimensional model that can predict a microrobot's trajectory and velocity over time with consideration of intermittent contact, non-point contact, and inclined or unstructured surfaces has yet to be developed.

In  this  paper,  we  present  techniques  for  simulating  motion of microrobots where there can be intermittent and non-point contact between the robot and the surface. The model we use is called a differential complementarity problem (DCP) model.
Let ${\bf u}\in \mathbb{R}^{n_1}$,  ${\bf v}\in \mathbb{R}^{n_2}$ and let ${\bf g}: \mathbb{R}^{n_1}\times \mathbb{R}^{n_2} \rightarrow \mathbb{R}^{n_1} $, ${\bf f}: \mathbb{R}^{n_1}\times \mathbb{R}^{n_2} \rightarrow \mathbb{R}^{n_2}$ be two vector functions and the notation $0 \le {\bf x} \perp {\bf y} \ge 0$ imply that ${\bf x}$ is orthogonal to ${\bf y}$ and each component of the vectors is non-negative. 
\begin{definition}
The differential (or dynamic) complementarity problem is to find $\bm{u}$ and $\bm{v}$ satisfying~\cite{Facchinei2007}:
$$\dot{\bf u} = {\bf g}({\bf u},{\bf v}), \ \ \ 0\le {\bf v} \perp {\bf f}({\bf u},{\bf v}) \ge 0 $$
\end{definition}
\begin{definition}
The mixed complementarity problem is to find $u$ and $v$ satisfying
$${\bf g}({\bf u},{\bf v})=0, \ \ \ 0\le {\bf v} \perp {\bf f}({\bf u},{\bf v}) \ge 0$$
\end{definition}
If the functions ${\bf f}$ and ${\bf g}$ are linear, the problem is called a mixed linear complementarity problem (MLCP). Otherwise, the problem is called a mixed nonlinear complementarity problem (MNCP). As we will discuss later, our discrete-time dynamics model is a MNCP.

Modeling the intermittent contact between bodies in motion as a complementarity constraint was first done by Lotstedt~\cite{Lotstedt82}. Subsequently, there was a substantial amount of effort in modeling and dynamic simulation with complementarity constraints~\cite{Moreau1988, AnitescuCP96, Pang1996, StewartT96, PfeifferG08, AcaryB2008, DrumwrightS12, Todorov14}. The DCP that models the equations of motion usually can not be solved in closed form. Therefore, a time-stepping scheme has been introduced to solve the DCP. Depending on the assumptions made when forming the discrete equation of motions, the discrete-time model can be divided into a mixed linear complementarity problem (MNCP)~\cite{AnitescuP97, AnitescuP02} and a mixed non-linear complementarity problem~\cite{Tzitzouris01,chakraborty2014geometrically}. Furthermore, depending on whether the distance function between the two bodies (which is a nonlinear function of the configuration) is approximated or linearized, the time-stepping scheme can also be further divided into geometrically explicit schemes~\cite{AnitescuCP96, StewartT96} and geometrically implicit schemes~\cite{Tzitzouris01,chakraborty2014geometrically,xie2016rigid}. 

All of the time stepping schemes mentioned above assume the contact between two objects to be point contact. However, at the microscale, the influence of adhesion and friction become more pronounced. Both of these factors scale with the surface contact area. \comment{Therefore, the dynamic model, which takes non-point contact into account and can handle surface-area dependent adhesive force, is needed. }Recently, we presented a dynamic model that takes non-point contact (where the contact mode could be point contact, line contact, or surface contact) into account~\cite{xie2016rigid}. The model belongs to a geometrically implicit time-stepping scheme, in which the distance function depends on the geometry and configurations of the rigid body. In this paper, we extend this model to handle surface area-dependent adhesive forces of a rigid body microrobot that will change during motion based on the contact mode. The resulting discrete time model is a MNCP problem.

There has been much effort to model and understand the effect of non-point frictional contact~\cite{Erdmann1994, Goyal1991, Howe1996}. We use the so called soft-finger contact model~\cite{MLS94} for the general dynamic simulation. The model is based on a maximum power dissipation principle and it assumes all the possible contact forces or moments should lie within an ellipsoid. At the microscale, adhesion is more pronounced and can have a significant effect on microrobot locomotion. It is the combined effect of forces that may stem from capillary effects, electrostatic charging, covalent bonding, hydrogen bonding, Casimir forces, or Van der Waals interactions~\cite{526162}. All of these forces, aside from forces arising from electrostatic charging, become negligible outside of the nanometer range. Van der Waals forces, for example, primarily act at ranges of 0.2-20 nm ~\cite{Diller2011a}. These forces can also be unpredictable and difficult to model individually. Therefore, we lumped the forces together into a single adhesion force and assume its effect is insignificant if there is no direct contact between the microrobot and the substrate. We formulated this adhesive force as an empirical relationship where it is proportional to the surface contact area. This relationship is useful because our dynamic model is capable of predicting the time-varying surface contact area. Electrostatic force is treated as a constant, since the distance between the microrobot and the substrate undergoes minimal change as the robot moves.

\section{Design Domain}

There is an increasingly vast set of materials and fabrication methods available for manufacturing complex structures at the microscale. In this paper, we investigate a small subset of these options that show potential for the purposes of tumbling magnetic microrobots. We consider two materials, SU-8 and PDMS (Polydimethylsiloxane), which are both polymers frequently used in microfluidics and MEMS applications. SU-8 is a negative photoresist that is sensitive to UV light and can form very rigid, high aspect ratio structures after becoming cross-linked ~\cite{arscott20148}. It shows exceptional biocompatibility and has been used for several biomedical applications such as cell encapsulation and neuronal probes ~\cite{nemani2013vitro}. Similarly, PDMS is a biocompatible silicone-based elastomer that cross-links to form solids with rubber-like consistency. It is the most commonly used material in the domain of experimental microfluidics, among other applications, due to its cost-effectiveness, excellent biocompatibility and permeability, low autofluorescence, and transparency ~\cite{raj2020pdms}. Both SU-8 and PDMS are still capable of cross-linking into solid structures after being doped with magnetic neodymium iron boron (NdFeB) particles, allowing tumbling microrobots to be formed from either material. Though difficult to machine using traditional manufacturing methods, thin films of SU-8 and PDMS can be processed into complex geometries with nanoscale resolution using photolithography or laser cutting processes. Both these fabrication methods, however, are limited to producing flat, two dimensional geometries and experience reduced performance when  incorporating doped magnetic particles.
\begin{figure}[!tbp]%
\centering
\includegraphics[width=\columnwidth]{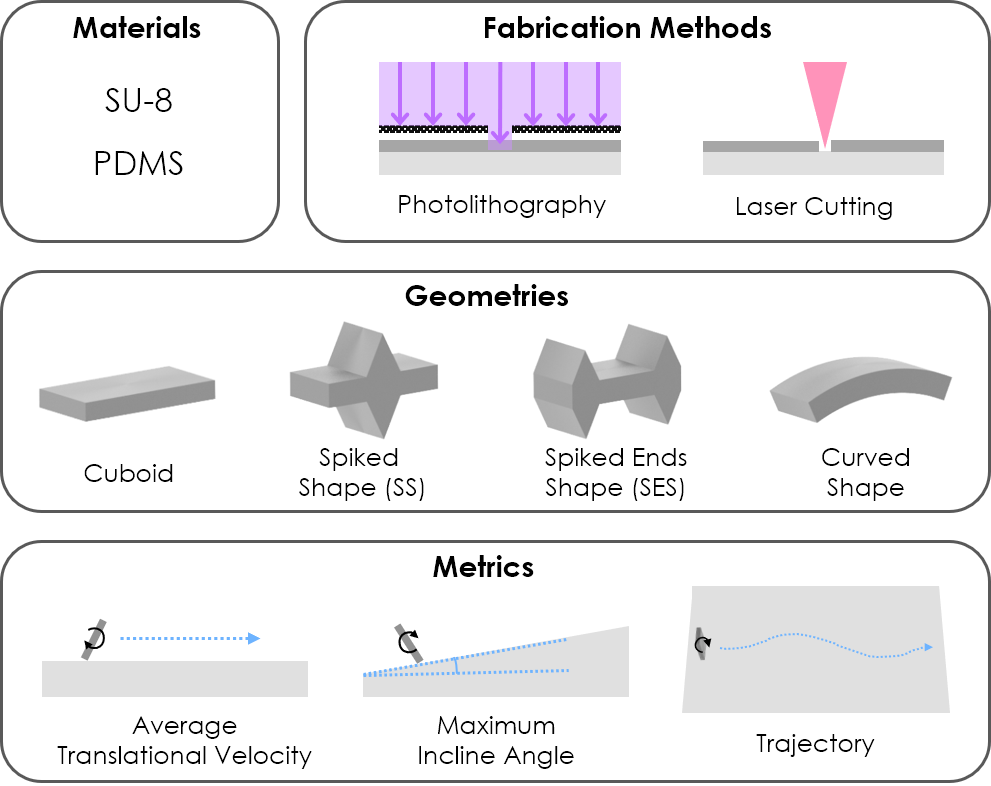}%
\caption{Overview of the tumbling microrobot design domain showing the materials, fabrication methods, and the geometries that we considered, as well as the metrics used to evaluate them.}
\label{design_domain} 
\end{figure}

The limitations set by these fabrication methods drive the shape space that is explored. In addition to the basic cuboid shape characterized in~\cite{Bi2018}, we consider alternative geometries for the tumbling microrobot, including a spiked shape (SS), a spiked ends shape (SES), and a curved shape. These new designs feature altered cross-sections and seek to improve microrobot responsiveness by minimizing area contact during the tumbling cycle and decreasing resistive adhesive forces. Our objective for the dynamic model is to use it for studying and predicting the best performing design without expending resources on iterative prototyping of physical microrobots. We also use it for understanding the robustness of the design to manufacturing errors. Metrics for evaluating degrees of microrobot performance include the average translational velocity in desired direction of motion, maximum climbable incline angle, and positional trajectory during the tumbling cycle. This data can all be captured optically using a digital microscope and furthered quantified using image processing. A high level overview of this paper's design domain is depicted in Figure~\ref{design_domain}.

\section{Dynamic Model for Rigid Body Systems}

In this section, we present an overview of the equations of motion of two rigid bodies in intermittent contact with each other. A microrobot moving on a surface may switch between having contact with the surface or no contact at all. Furthermore, when the robot is in contact, the contact may be a sliding or sticking contact (i.e., no relative velocity between the points on the objects in the contact region). Depending on the geometry of the robot and its configuration, the contact mode may also be point contact, line contact, or surface contact.  A key requirement for building dynamic simulators for the microrobots is the ability to handle surface area-dependent adhesive forces that will change during motion based on the contact mode. We will therefore use a complementarity-based model of dynamics that can handle the transition between no-contact and contact as well as sticking and sliding contact in a unified manner. Furthermore, since we can have non-point contact, we will use the equations of motion in~\cite{xie2016rigid} as our basic model for the dynamics. 

When fabricating the microrobots manufacturing errors inevitably exist. It is possible that these errors in geometry and/or magnetization axis alignment may cause the microrobots to tilt or flip unexpectedly during the tumbling motion. Thus, the motion of the microrobot is not restricted to a two-dimensional plane and the three-dimensional dynamic model in~\cite{xie2016rigid} is necessary to simulate the dynamics of the microrobots.

\begin{figure}
\centering
\begin{subfigure}[b]{0.8\columnwidth}
\includegraphics[width=\columnwidth]{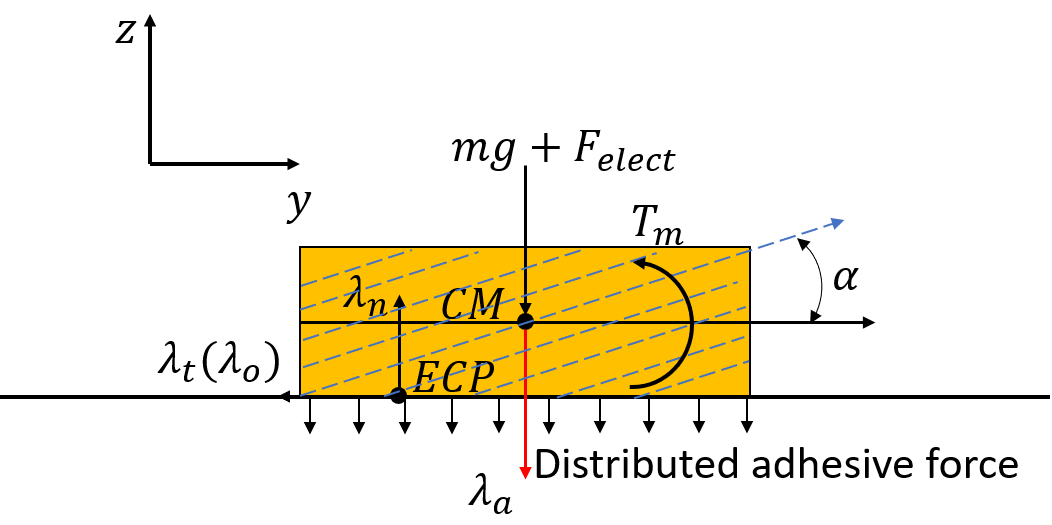}%
\caption{}
\label{figure_surface_contact} 
\end{subfigure}
\begin{subfigure}[b]{0.8\columnwidth}
\includegraphics[width=\columnwidth]{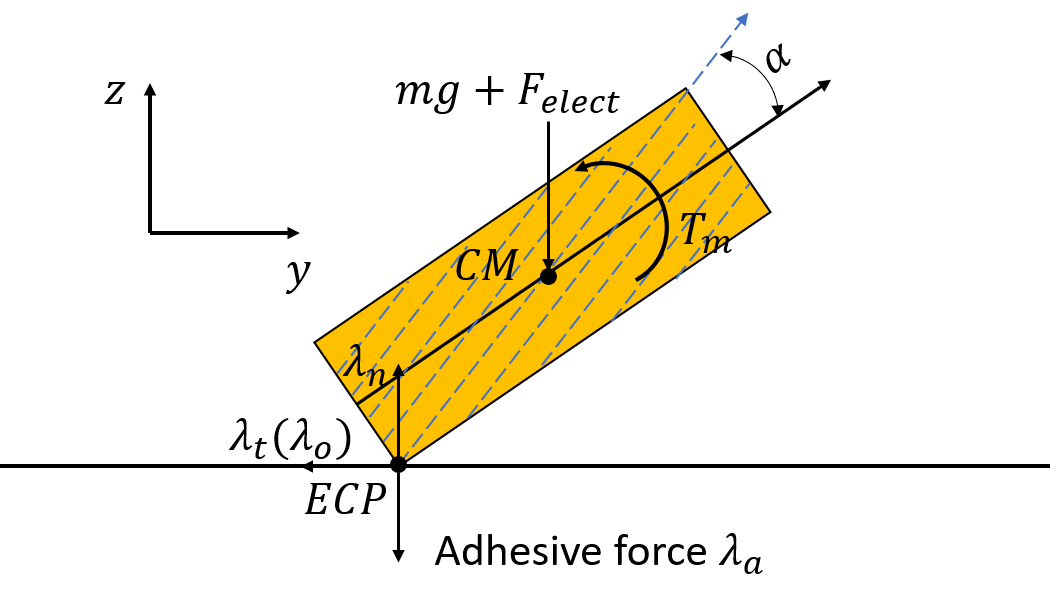}%
\caption{}
\label{figure_line_contact} 
\end{subfigure}
\caption{Force diagrams in 2D when robot has (a) surface contact and (b) line contact with horizontal surface in 2D. The dashed lines in blue represent the internal magnetic alignment. The adhesive force is distributed uniformly over the surface area. When robot has line contact with the surface, the adhesive force is almost zero. }
\label{Contact_2D}
\end{figure}

The general equations of motion has three key parts: (i) Newton-Euler  differential equations of motion giving state update, (ii) algebraic and complementarity constraints modeling the fact that two rigid bodies cannot penetrate each other, and (iii) model of the contact force and moments acting on the contact patch. For general rigid body motion, the model of contact forces and moments use Coulomb's assumption that {\em the normal force acting between two objects is independent of the nominal contact area between the two objects}. This is a reasonable assumption for nominally rigid objects at macroscopic length scales, where the inertial forces are dominating. However, at the length-scale of microrobots, the force of adhesion between the contacting surfaces is comparable to inertial forces. So, the contact model should also take into consideration the effect of the surface-area dependent forces. These forces, combined under a single adhesive force, are illustrated in Figure~\ref{Contact_2D}.

For simplicity of exposition, we assume one body to be static. 
Let ${\bf V} = [{\bf v}^T ~\bm{\omega}^T]^T$ be the generalized velocity of the rigid body, where ${\bf v} \in \mathbb{R}^3$ is the linear velocity and $\bm{\omega} \in \mathbb{R}^3$ is the angular velocity of the rigid body. Let ${\bf q}$ be the configuration of the rigid body, which is a concatenated vector of the position and a parameterization of the orientation of the rigid body. 

{\bf Newton-Euler Equations of Motion}:
The Newton-Euler equations of motion of the rigid body are:
\begin{equation} \label{eq_general}
{\bf M}({\bf q})
{\dot{\bm{V}}} = 
{\bf W}_{n}\lambda_{n}+
{\bf W}_{t}\lambda_{t}+
{\bf W}_{o}\lambda_{o}+
{\bf W}_{r}\lambda_{r}+
\bm{\lambda}_{app}+\bm{\lambda}_{vp}
\end{equation}
where ${\bf M}({\bf q})$ is the inertia tensor, $\bm{\lambda}_{app}$ is the vector of external forces and moments (including gravity), $\bm{\lambda}_{vp}$ is the  centripetal and Coriolis forces. The magnitude of the normal contact force is $\lambda_n$. The magnitude of tangential contact forces are $\lambda_t$ and $\lambda_o$. The magnitude of the moment due to the tangential contact forces about the contact normal is $\lambda_r$. The vectors ${\bf W}_n$, ${\bf W}_t$, ${\bf W}_o$ and ${\bf W}_r$ map the contact forces and moments from the contact point to the center of mass of the robot. The expressions of ${\bf W}_n$, ${\bf W}_t$, ${\bf W}_o$ and ${\bf W}_r$ are:
\begin{equation}
\begin{aligned}
\label{equation:wrenches}
{\bf W}_{n} =  \left [ \begin{matrix} 
{\bf n}\\
{\bf r}\times {\bf n}
\end{matrix}\right],
{\bf W}_{t} =  \left [ \begin{matrix} 
{\bf t}\\
{\bf r}\times {\bf t}
\end{matrix}\right],
{\bf W}_{o} =  \left [ \begin{matrix} 
{\bf o}\\
{\bf r}\times {\bf o}
\end{matrix}\right],
{\bf W}_{r} =  \left [ \begin{matrix} 
{\bf 0}\\
\ \ \bm{n} \ \
\end{matrix}\right]
\end{aligned}
\end{equation}
where $({\bf n},{\bf t},{\bf o}) \in \mathbb{R}^3$ are the axes of the contact frame, ${\bf 0} \in \mathbb{R}^3$ is a column vector with each entry equal to zero. As shown in Figure~\ref{Contact_2D}, vector ${\bf r} = [a_x-q_x,a_y-q_y,a_z-q_z]$ is the vector from equivalent contact point (ECP) ${\bf a}$, to center of mass (CM), where $(q_x, q_y, q_z)$ is the position of the CM. In the next section, we will provide definition for the equivalent contact point (ECP). Please note that Equation~\eqref{eq_general} is a system of $6$ differential equations.

{\bf Modeling Rigid Body Contact Constraints:} The contact model that we use is a complementarity-based contact model as described in~\cite{xie2016rigid,chakraborty2014geometrically}. In~\cite{xie2016rigid}, we introduced the notion of an equivalent contact point (ECP) to model non-point contact between objects. 
\begin{definition}
 Equivalent Contact Point (ECP) is a unique point on the contact surface that can be used to model the surface (line) contact as point contact  where  the  integral of the total moment (about the point) due to the distributed normal force on the contact patch is zero.
\end{definition}
 The ECP defined here is the same as the center of friction. Now let's describe the contact model mathematically. Let two objects $F$ and $G$ be defined by intersection of convex inequalities $f_{i}(\bm{\zeta}_1)\le 0, i = 1,..,m$, and $g_{j}(\bm{\zeta}_2)\le 0, j = m+1,..,n$ respectively. Let ${\bf a}_1$ and ${\bf a}_2$ be pair of ECPs or closest points (when objects are separate) on F and G, respectively. The complementarity conditions for nonpenetration can be written as either one of the following two sets of conditions~\cite{chakraborty2014geometrically}:
\begin{equation}
\begin{aligned}
\label{equation:contact_multiple_comp}
0 \le \lambda_{n} \perp \mathop{max}_{1,...,m} f_{i}({\bf a}_2) \ge 0\\
0 \le \lambda_{n} \perp \mathop{max}_{j=m+1,...,n}g_{j}({\bf a}_1) \ge 0
\end{aligned}
\end{equation}

The solution of ECP's ${\bf a}_1$ and ${\bf a}_2$ is given by the following minimization problem:
\begin{equation}
\label{equation:optimazation}
({\bf a}_1,{\bf a}_2) = arg \min_{\bm{\zeta}_1,\bm{\zeta}_2}\{ \|\bm{\zeta}_1-\bm{\zeta}_2 \| \ f_{i}(\bm{\zeta}_1) \le 0,\ g_{j}(\bm{\zeta}_2) \le 0 \}
\end{equation}
where $i =1,...,m$ and $j=m+1,...,n$.

Using a slight modification of the KKT conditions for the optimization problem in Equation~\eqref{equation:optimazation}, and combining it with either one of the conditions in Equation~\eqref{equation:contact_multiple_comp}, we get the complete contact model between two rigid bodies:
\begin{equation}
\begin{aligned}
\label{equation:re_contact_multiple1}
&{\bf a}_{1}-{\bf a}_{2} = -l_{k}\mathcal{C}({\bf F},{\bf a}_1), \
\mathcal{C}({\bf F},{\bf a}_1)= -\mathcal{C}({\bf G},{\bf a}_2)\\
0 \le &\left[ \begin{matrix} l_{i}\\ l_{j}\\ \lambda_{n}  \end{matrix} \right] \perp \left[ \begin{matrix} &-f_{i}({\bf a}_{1}),  \\ &-g_{j}({\bf a}_{2}), \\ &\max\limits_j f_{j}({\bf a}_2)  \end{matrix} \right] \ge 0 \\
&i = 1,...,m, \quad j = m+1,...,n.
\end{aligned}
\end{equation}
where $k$ is the index of active constraint on body $F$, and the normal cones are: $\mathcal{C}({\bf F},{\bf a}_1) = \nabla f_{k}({\bf a}_{1})+\sum_{i = 1,i\neq k}^m l_{i}\nabla f_{i}({\bf a}_{1})$, $\mathcal{C}({\bf G},{\bf a}_2) = \sum_{j = m+1}^n l_{j} \nabla g_{j} ({\bf a}_{2})$.

\begin{figure}
\centering
\includegraphics[width=0.8\columnwidth]{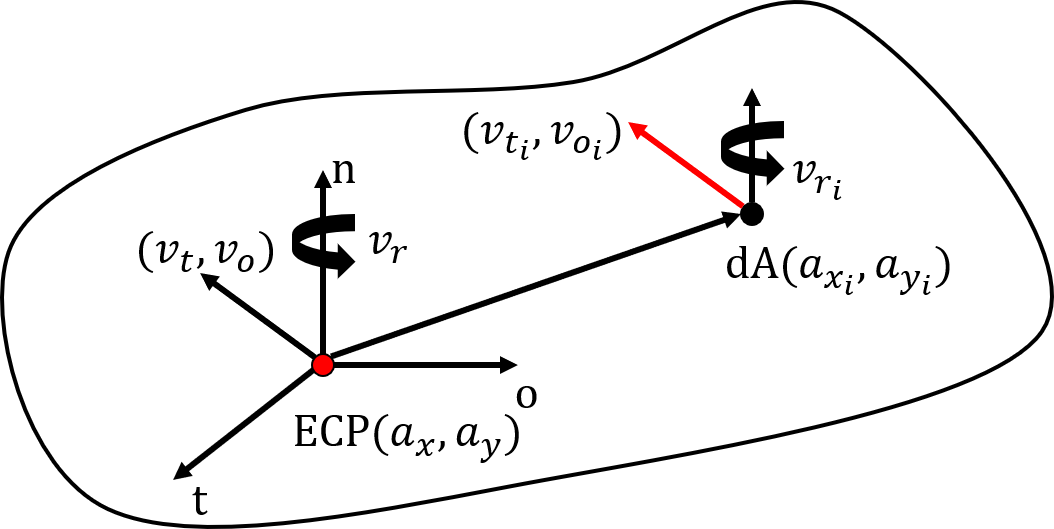}
\caption{Notation for planar sliding motion. } 
\label{figure_pressure} 
\end{figure} 

{\bf Friction Model:} We use a friction model based on the maximum power dissipation principle, which has been previously proposed in the literature for point contact~\cite{Moreau1988}. The maximum power dissipation principle states that among all the possible contact forces and moments that lie within the friction ellipsoid, the forces that maximize the power dissipation at the contact are selected. For non-point contact, we will use a generalization of the maximum power dissipation principle, where, we select contact forces/moments and contact velocities that maximize the power dissipation over the entire contact patch. We will now show that the problem formulation using the power loss over the whole contact patch can be reduced to the friction model for point contact with the ECP as the chosen point. Mathematically, the power dissipated over the entire surface, $P_c$ is 
\begin{equation}
\label{eq:power}
P_c = -\int_{A} (v_{ti}\beta_{ti}+v_{oi}\beta_{oi}+v_{ri}\beta_{ri})dA
\end{equation}
where $v_{ti},v_{oi}$ are the linear sliding velocities and $v_{ri}$ is the angular velocity at $dA$, $\beta_{ti},\beta_{oi}$ are the frictional force per unit area and $\beta_{ri}$ is the resistive moment per unit area at $dA$, about the normal to the contact patch. We will assume a planar contact patch which implies that the contact normal is the same at all points on the contact patch. As shown in Figure~\ref{figure_pressure}, the angular velocity is constant across the patch, i.e., $v_{ri} = v_r$, for all $i$. Let $v_t$ and $v_o$ be the components of tangential velocities at the ECP. From basic kinematics, we know that $v_{ti} = v_t - v_{ri}(a_{yi}-a_y)$ and $v_{oi} = v_o + v_{ri}(a_{xi}-a_x)$, where ($a_x$, $a_y$) are the $x$ and $y$ coordinates of the ECP and ($a_{xi}$, $a_{yi}$) are the $x$ and $y$ coordinates of a point on the patch. Substituting the above in Equation~\eqref{eq:power} and simplifying, we obtain
\begin{equation}
\label{eq:power2}
P_c = -\left[\int_{A} v_{t}\beta_{ti}dA + \int_{A}v_{o}\beta_{oi}dA +\int_{A}v_{ri}\beta_{ri}^{\prime}dA \right ]
\end{equation}
where $\beta^{\prime}_{ri} = \beta_{ri} -\beta_{ti}(a_{yi}-a_y) + \beta_{oi}(a_{xi}-a_x)$.
%
%
By noting that $\int \beta_{ti}dA = \lambda_t, \int \beta_{oi}dA  = \lambda_o, \int \beta^{\prime}_{ri}dA  = \lambda_r$, where $\lambda_t$, $\lambda_o$ are the net tangential forces at the ECP and $\lambda_r$ is the net moment about the axis normal to the contact patch and passing through the ECP, the power dissipation over the entire contact patch is given by $P_c = - (v_t \lambda_t + v_o\lambda_o + v_r \lambda_r)$. For specifying a friction model, we also need a law or relationship that bounds the magnitude of the friction forces and moments in terms of the magnitude of the normal force~\cite{goyal1991planar}. Here, we use an ellipsoidal model for bounding the magnitude of tangential friction force and friction moment. This friction model has been previously proposed in the literature~\cite{goyal1991planar,Moreau1988,xie2016rigid,chakraborty2014geometrically} and has some experimental justification~\cite{howe1996practical}.
Thus, the contact wrench is the solution of the following optimization problem: 
\begin{equation}
\begin{aligned}
\label{equation:friction}
{\rm max} \quad -(v_t \lambda_t + v_o\lambda_o + v_r \lambda_r)\\
{\rm s.t.} \quad \left(\frac{\lambda_t}{e_t}\right)^2 + \left(\frac{\lambda_o}{e_o}\right)^2+\left(\frac{\lambda_r}{e_r}\right)^2 - \mu^2 \lambda_n^2 \le 0
\end{aligned}
\end{equation}
where the magnitude of contact force and moment at the ECP, namely, $\lambda_t$, $\lambda_o$, and $\lambda_r$ are the optimization variables. The parameters, $e_t$, $e_o$, and $e_r$ are positive constants defining the friction ellipsoid and $\mu$ is the coefficient of friction at the contact~\cite{howe1996practical,trinkle1997dynamic}.
As stated before, we use the contact wrench at the ECP to model the effect of entire distributed contact patch.\comment{ Therefore $v_t$ and $v_o$ are the tangential components of velocity at the ECP; $v_r$ is the relative angular velocity about the normal at ECP. } 
Note that there is {\em no assumption made on the nature of the pressure distribution between the two surfaces}. A key aspect of this work, which is different from previous effort, is that, here we consider that the normal force can be a function of the contact surface area. We will elaborate on how this is done within the context of the discrete-time framework, since this requires that we identify the contact surface as part of our dynamic simulation algorithm.  

Using the Fritz-John optimality conditions of Equation~\eqref{equation:friction}, we can write~\cite{trinkle2001dynamic}:
\begin{equation}
\begin{aligned}
\label{eq:friction}
0&=
e^{2}_{t}\mu \lambda_{n} 
v_t+
\lambda_{t}\sigma \\
0&=
e^{2}_{o}\mu \lambda_{n}  
v_o+\lambda_{o}\sigma \\
0&=
e^{2}_{r}\mu \lambda_{n}v_r+\lambda_{r}\sigma\\
0& \le \mu^2\lambda_n^2- \lambda_{t}^2/e^{2}_{t}- \lambda_{o}^2/e^{2}_{o}- \lambda_{r}^2/e^{2}_{r} \perp \sigma \ge 0
\end{aligned}
\end{equation}
where $\sigma$ is a Lagrange multiplier corresponding to the inequality constraint in~\eqref{equation:friction}. 
\section{Modeling for Tumbling Microrobots}

As shown in Figure~\ref{figure_robot}, the magnetic microscale tumbling robot ($\mu$TUM) presented in this paper is cuboid-shaped and embedded with magnetic particles. The robot's magnetic features are aligned along a certain direction and optimally it should to be aligned along lengthwise direction of the robot. An alignment offset angle is defined when there exists an angular difference between actual alignment direction and the desired alignment direction.

There exists one magnetic field which rotates counterclockwise about the $x$ axis of the world frame. When the magnetic alignment of the field differs from that of the robot, a magnetic torque is applied on the robot until it is realigned with the field. Therefore, a rotating magnetic field causes the robot to rotate about the same axis. As shown in Figure~\ref{figure_robot}, if the robot is resting on the surface, the rotating field causes the tumbling motion of the robot, i.e., the robot will move forward by continuously flipping end-over-end.

{\bf Notation}:
The following notation will be used for defining the problem mathematically:

\begin{itemize}
\item[$\circ$] $L, M, H$ $-$  length, width and height of the robot
\item[$\circ$] ${\bf M} = diag(m,m,m,I_{xx},I_{yy},I_{zz})$ $-$ inertia tensor of the robot, where $m$ represents mass and $I_{(.)}$ represents moment of inertia in body frame
\item[$\circ$] $F_{elect}$ $-$ electrostatic force of the robot
\item[$\circ$] $V_m$ $-$ magnetic volume of the robot
\item[$\circ$] ${\bf E}\in \mathbb{R}^3$ $-$ magnetization of the robot (The blue dashed lines in Figure~\ref{Contact_2D})
\item[$\circ$] $\alpha$ $-$ magnetic alignment offset angle
\item[$\circ$] ${\bf B}\in \mathbb{R}^3$, ${\bf T}_m \in \mathbb{R}^3$ $-$ magnetic field strength and torque
\item[$\circ$] $f_{rot}$ $-$ the frequency of rotational field 
\item[$\circ$] $\mu$ $-$ friction coefficient between robot and surface
\item[$\circ$] $C$ $-$  coefficient of adhesive force between robot and surface
\item[$\circ$] $A_{contact}$ $-$ area of contact region between robot and surface
\item[$\circ$] $\lambda_{a}$ $-$ the adhesive force between robot and surface
\item[$\circ$] $e_t, e_o, e_r$ $-$ friction parameters defining the friction ellipsoid
\item[$\circ$] ${\bf n}\in \mathbb{R}^3$ $-$ the contact normal, which is used to define the normal axis of contact frame
\item[$\circ$] ${\bf t}\in \mathbb{R}^3$, ${\bf o}\in \mathbb{R}^3$ $-$ tangential axies of the contact frame
\item [$\circ$] ${\bf v} = [v_x, v_y, v_z]$ $-$ linear velocity of the robot
\item [$\circ$] ${\bf w} = [w_x, w_y, w_z]$ $-$ angular velocity of the robot
\item [$\circ$] $\lambda_n, \lambda_t, \lambda_o$  $-$ normal and tangential contact forces
\item [$\circ$] $\lambda_r$ $-$ frictional moment about contact normal ${\bf n}$
\item [$\circ$] ${\bf a}_1 \in \mathbb{R}^3, {\bf a}_2 \in \mathbb{R}^3$ $-$ pair of equivalent contact points (ECP)
\item [$\circ$] $\sigma$ $-$ Lagrange multiplier associated with the friction model, which represents the magnitude of slip velocity
\item [$\circ$] ${\bf l}_1 =[l_1,...,l_m], {\bf l}_2 =[l_{m+1},...,l_n]$ $-$ Lagrange multipliers in contact constraints
\end{itemize}
The magnetic torque ${\bf T}_m$ applied to the microrobot is:
\begin{equation}
\label{Torque}
    {\bf T}_m = V_m{\bf E}\times{\bf B}
\end{equation}
The direction of the adhesive force between the robot and the surface, $\lambda_{a}$, is along the negative direction of the contact normal, ${\bf n}$, and its value depends on the material of the object and the area of contact region. The expression for $\lambda_{a}$ is:
\begin{equation}
\label{ADH}
\lambda_{a} = CA_{contact}  
\end{equation}

{\bf Newton-Euler Equations for Tumbling Microrobot}: 
As shown in Figure~\ref{Contact_2D}, the generalized applied force $\bm{\lambda}_{app} \in \mathbb{R}^6$ acting on CM of the robot includes gravity force $mg$, electrostatic force $F_{elect}$, adhesive force $\lambda_a$ and magnetic torque ${\bf T}_m \in \mathbb{R}^3$. The contact wrench acting on the ECP includes normal contact force, $\lambda_n$, and frictional forces and moments, $\lambda_t, \lambda_o$ and $\lambda_r$. The generalized velocity is ${\bf V} = [{\bf v},{\bf w}]$. The Newton-Euler equations are:
\begin{equation}
\label{NE}
{\bf M} \dot{\bm{\nu}} = {\bf W}
\left[\begin{matrix}
\lambda_{n}-\lambda_a\\
\lambda_{t}\\
\lambda_{o}\\
\lambda_{r}
\end{matrix}
\right]
+
\left [\begin{matrix}
0\\
0\\
-(mg+F_{elect})\\
{\bf T}_m
\end{matrix} \right]
+\bm{\lambda}_{vp}
\end{equation}
where the mapping matrix ${\bf W} = [{\bf W}_{n}, {\bf W}_{t}, {\bf W}_{o}, {\bf W}_{r}] \in \mathbb{R}^{6\times 4}$ is computable based on Equation~\eqref{equation:wrenches}. The magnetic torque ${\bf T}_m$ is based on Equation~\eqref{Torque}. The adhesive force $\lambda_a$ is in the opposite direction of the normal force $\lambda_n$, and it's value is computed by the Equation~\eqref{ADH}. Please note that Equation~\eqref{NE} is a system of 6 differential equations.

{\bf Discrete-time dynamic model}:We use a velocity-level formulation and an Euler time-stepping scheme to discretize the above system of equations. Let superscripts $u$ be the beginning of current time step, $u+1$ be the end of current time step, and $h$ be the time step length. Let $\dot{\bf V}  \approx ({\bf V}^{u+1} -{\bf V}^u)/h$ and impulse $p_{(.)} = h\lambda_{(.)}$, we get the following discrete-time system. The system of equations in general is a mixed nonlinear complementarity problem. The vector of unknowns, ${\bf z}$, can be partitioned into ${\bf z} = [{\bf u}_z,{\bf v}_z]$, where:
\begin{equation*}
    {\bf u}_z = [{\bf V};{\bf a}_1;{\bf a}_2;p_t;p_o;p_r], \ {\bf v}_z = [{\bf l}_1;{\bf l}_2;\sigma;p_n]
\end{equation*}

The equality constraints in the mixed NCP are:
\begin{equation}
\begin{aligned}
&{\bf M}^u ({\bf V}^{u+1}- {\bf V}^{u}) =  {\bf W}^{u+1}
\left[\begin{matrix}
p^{u+1}_{n}-p^{u}_a\\
p^{u+1}_{t}\\
p^{u+1}_{o}\\
p^{u+1}_{r}
\end{matrix}
\right]
-
\left [\begin{matrix}
0\\
0\\
mgh+p_{elect}\\
-{\bf T}^u_mh
\end{matrix} \right]
 - {\bf p}_{vp}^{u} \\
&0 = {\bf a}^{u+1}_1-{\bf a}^{ u+1}_2+l^{u+1}_{k} \mathcal{C}({\bf F},{\bf a}^{u+1}_{1})\\
&0 = \mathcal{C}({\bf F},{\bf a}^{u+1}_{1})+\mathcal{C}({\bf G},{\bf a}^{  u+1}_{2})\\
&0 = \mu e_t^2p^{u+1}_{n}{{\bf W}^{T u+1}_{t}}{\bf V}^{u+1}+p^{u+1}_{t}\sigma^{u+1}\\
&0 = \mu e_o^2p^{u+1}_{n}{\bf W}^{T u+1}_{o}{\bf V}^{u+1}+p^{u+1}_{o}\sigma^{u+1}\\
&0 = \mu e_r^2p^{u+1}_{n}{\bf W}^{T u+1}_{r}{\bf V}^{u+1}+p^{u+1}_{r}\sigma^{u+1}
\end{aligned}
\end{equation}
The complementarity constraints on ${\bf v}_z$ are:
\begin{equation}
\begin{aligned}
0 \le &\left[ \begin{matrix} {\bf l}^{u+1}_{1}\\ {\bf l}^{u+1}_{2}\\  \sigma^{u+1} \\p^{u+1}_{n}  \end{matrix} \right] \perp \left[ \begin{matrix} -{\bf f}({\bf a}^{u+1}_{1}) \\-{\bf g}({\bf a}^{ ^{u+1}}_{2})\\  \xi  \\ \max {\bf f}({\bf a}^{ ^{u+1}}_2) \\
\end{matrix} \right] \ge 0 
\end{aligned}
\end{equation}
where $\xi =(\mu p^{u+1}_{n})^2-(p^{u+1}_{t}/e_t)^2 -(p^{u+1}_{o}/e_o)^2 -(p^{u+1}_{r}/e_r)^2$. Furthermore, the adhesive impulse $p^u_{a}$ is required as input at the beginning of each time step. We can compute $p^u_{a}$ based on Equation~\eqref{ADH}. However, in order to compute $p^u_{a}$, we need to know the contact are at each time step. However, this is not part of our solution to the dynamic model. In next section, we will discuss the procedure to compute the contact area, $A_{contact}$.

{\bf Computing the area of contact region}:
In general, the area of contact region, $A_{contact}$, depends on the geometry and configurations of objects in contact, which is hard to describe mathematically. However, in our case, the contact happens between the microrobot ($\mu$TUM) and the planar surface. The contact region is the side of the robot in contact with the surface. The geometry and dimension of the robot can be measured a priori and we can compute the area of each side of the robot. The next question is: which side of the robot is in contact at the current time?

The question can be answered by utilizing Lagrange multipliers of contact constraints. Based on the complementary condition, once  $l^{u+1}_i > 0$,  its associated constraint $f_{i}({\bf a}^{u+1}_{1}) =0$, i,e., the Equivalent contact point should lie on the constraint or side $i$. If $p^{u+1}_n > 0$, which indicates robot has contact on the surface at the end of the current time, the active constraint or side $i$ will be the side of robot that has contact with the surface. 

To sum up, first we can compute the area of each side of the robot based on the knowledge of robot's geometry and dimensions. Then, we solve the discrete-time model at each time step. The solutions for $l^{u+1}_i$ and $p_n^{u+1}$ will be utilized to identify the side or boundary of the robot on contact and return us $A^{u+1}_{contact}$. Eventually, based on Equation~\eqref{ADH}, we compute adhesive impulse $p^{u+1}_a$, which would be used as input for next time step. In the subsequent  sections, we will utilize the method in the simulation to estimate the effect of adhesion.
\section{Dynamic Model Validation}
To validate our dynamic model, we compare our simulation results to experimental results. We use the following experimental tests: (i) tumbling locomotion tests on paper (ii) inclined plane traversal tests on paper and (iii) inclined plane traversal tests on aluminium. 
In the tumbling locomotion tests, the $\mu$TUM moves on a flat horizontal surface and we use the average translational speed, $v$, in the desired direction of motion as the metric. In inclined tests, we measure the maximum climbable incline angle, $\varphi$, of the microrobot. We perform the tests in the simulation and validate the results with experiments. We will first discuss the experimental setup and then discuss the results of the three tests.

\subsection{Experimental Setup}
The microrobots used in the experiments described in this section are composed of two SU-8 polymer ends doped with magnetic NdFeB particles and a non-magnetic middle section that is entirely made up of SU-8 polymer. Their external dimensions are: Length, $L = 0.8\times 10^{-3}m$, Width, $W =0.4\times 10^{-3}m$, and Height, $H = 0.1\times 10^{-3}m$. The microrobots were fabricated using a two-step photolithography process described in \cite{Bi2018}. Two different generations of microrobots were used in the experiments. The material properties for the first generation are listed in Table~\ref{table_1} and they were used for the experiments on paper.  The material properties for the second generation are listed in Table~\ref{table_3} and they were used for the experiments on aluminium.   
The second generation robots underwent an additional step where they were exposed to a 9 T uniform magnetic field generated by a PPMS machine (Quantum Design) after the SU-8 curing process. This field was strong enough to realign the embedded NdFeB particles homogeneously within the cured SU-8 and the resulting magnetization was approximately three times larger than earlier tumbling microrobot iterations. 

\begin{figure}
\centering
\includegraphics[width=3in]{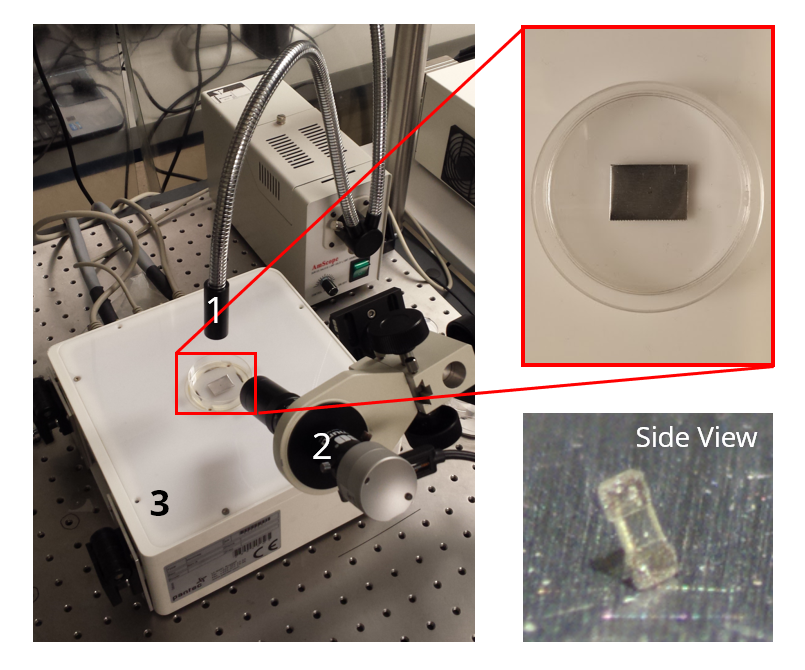}
\caption{Experimental setup with halogen lamp (1), side camera (2) and MFG-100 system (3). Additional images show an aluminum surface inside a petri dish at the center of the workspace and a side view of the {$\mu$}TUM as seen through the camera.}
\label{figure:Experimental_Setup} 
\vspace{-0.25in}
\end{figure}

A system of eight electromagnetic coils (MFG-100 system, MagnetibotiX AG) was used to generate the rotating magnetic field that actuates the microrobots. Figure~\ref{figure:Experimental_Setup} depicts the experimental setup. While the microrobots used for the experiments have three distinct sections, our simulation simplifies them into single, homogeneous blocks of uniform mass distribution. We argue this assumption is acceptable at the microscale, where factors such as weight and inertia are much smaller in magnitude than factors proportional to distance and surface area, such as adhesion and electrostatic forces. 

Several properties utilized in the simulation were derived from physical measurements of related parameters. To obtain the adhesion coefficient for the substrate of interest, the microrobot was laid flat over the substrate in dry air. The external magnetic field was set to a static vertical orientation and the field strength was incrementally increased from zero until the microrobot started rotating upwards. The field strength at which rotation occurred was used to calculate the magnetic torque that counteracted the adhesion force resisting upwards motion. Dividing this torque by the moment arm and by the total contact surface area of the robot resulted in the adhesion coefficient for that substrate. To estimate the friction coefficient, a wafer of SU-8 was placed over a sheet of the substrate of interest in dry air. The SU-8 side of the wafer was placed in contact with the substrate and 20 grams of additional mass was attached to the other side, ensuring that the dominant force between the microrobot and the substrate would be weight instead of adhesion or electrostatic forces. The substrate was then tilted from a horizontal position until the wafer started slipping downwards. The angle at which slippage occurred was noted and the friction coefficient for the substrate was  approximated by taking the tangent of this angle.


\subsection{Tumbling Locomotion Tests on Paper}
\begin{table}
\caption{Parameters for $\mu$TUM on paper.}
\begin{tabular}{c c c}
\hline \hline
Description   & Value & Units \\
\hline
Mass (m)  & $3.78\times 10^{-8}$  & kg \\ 
Electrostatic Force ($F_{elect}$) & $6.54\times 10^{-7}$  & N\\   
Friction Coefficient ($\mu$) & $0.3$  & -\\
Magnetic Alignment Offset  ($\alpha$) & $27$  & degree\\
Magnetic Volume  ($V_m$) & $2.9\times 10^{-11}$  & $m^3$\\
Magnetization ($|{\bf E}|$) & $15000$  & $A/m$\\
Coefficient of adhesion force ($C$) & 1.19 &$N/m^2$ \\ 
\hline
\end{tabular}
\label{table_1}
\end{table}
The first scenario investigated is for tumbling locomotion of the $\mu$TUM traversing a dry paper substrate. The parameters are, again, listed in Table~\ref{table_1}. As listed in the table, the coefficient of adhesion force is $C = 1.19 N/m^2$. Thus, the effect of adhesion between the robot and paper can not be ignored. Therefore, in our simulation, we take the adhesive force into account and compute the value based on our dynamic model. The simulation is performed in order to evaluate the robot's performance on the substrate under varying field rotation frequencies. If the robot tumbles without slipping on the rough paper surface, the robot's average translational speed, $v$, should be approximately equal to two times the sum of body length and body height $(L+H)$ multiplied by the field rotational frequency $f_{rot}$~\cite{Bi2018}:
\begin{equation}
\label{model_non_slip}
    v = 2(L+H)f_{rot}
\end{equation}
\noindent
\begin{figure}
\centering
\includegraphics[width=2.5in]{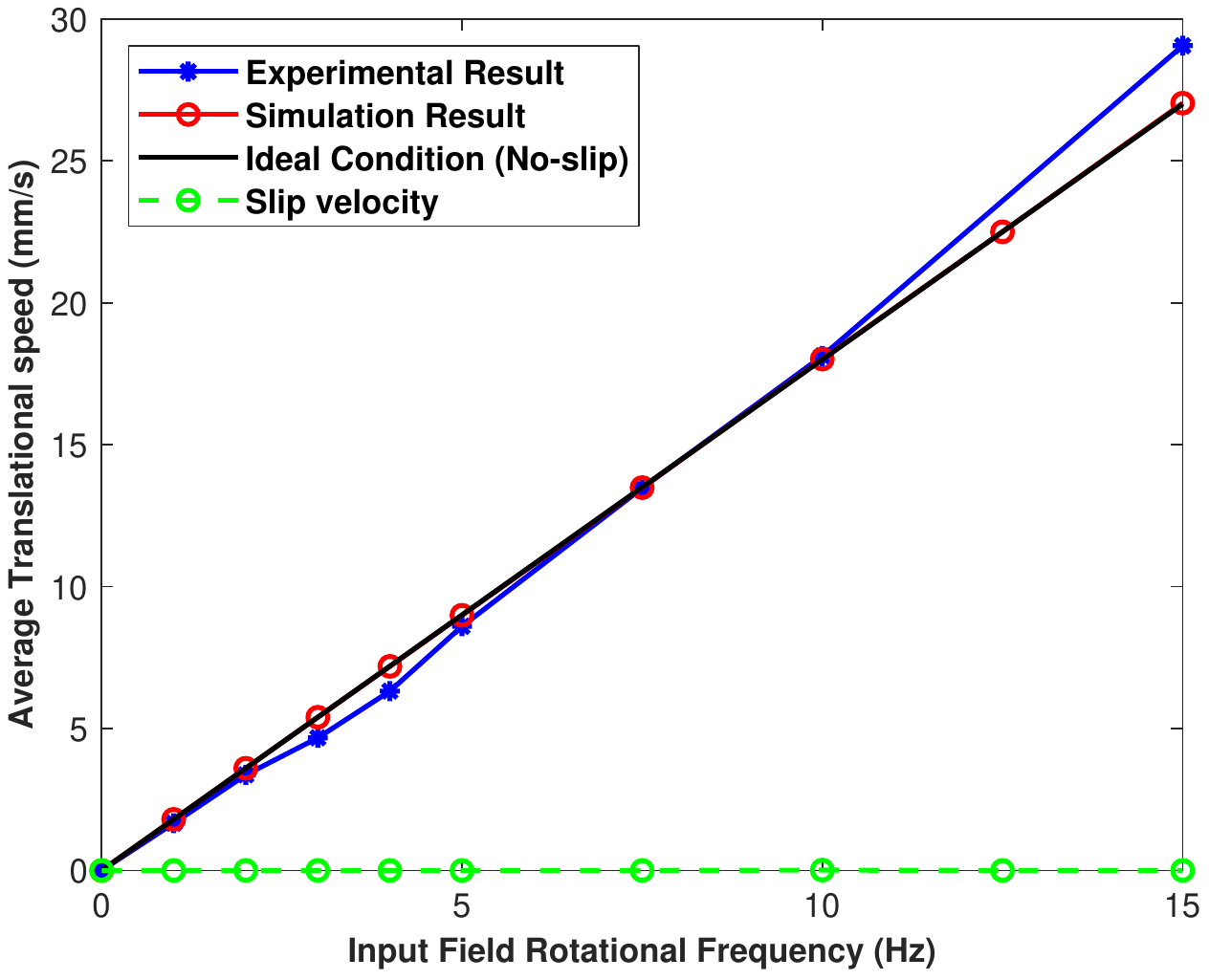}
\caption{Tumbling locomotion tests on paper ( $10mT$ field).}
\label{figure:Tumbling} 
\end{figure} 
\begin{figure}[!htbp]
\centering
\includegraphics[width=2.5in]{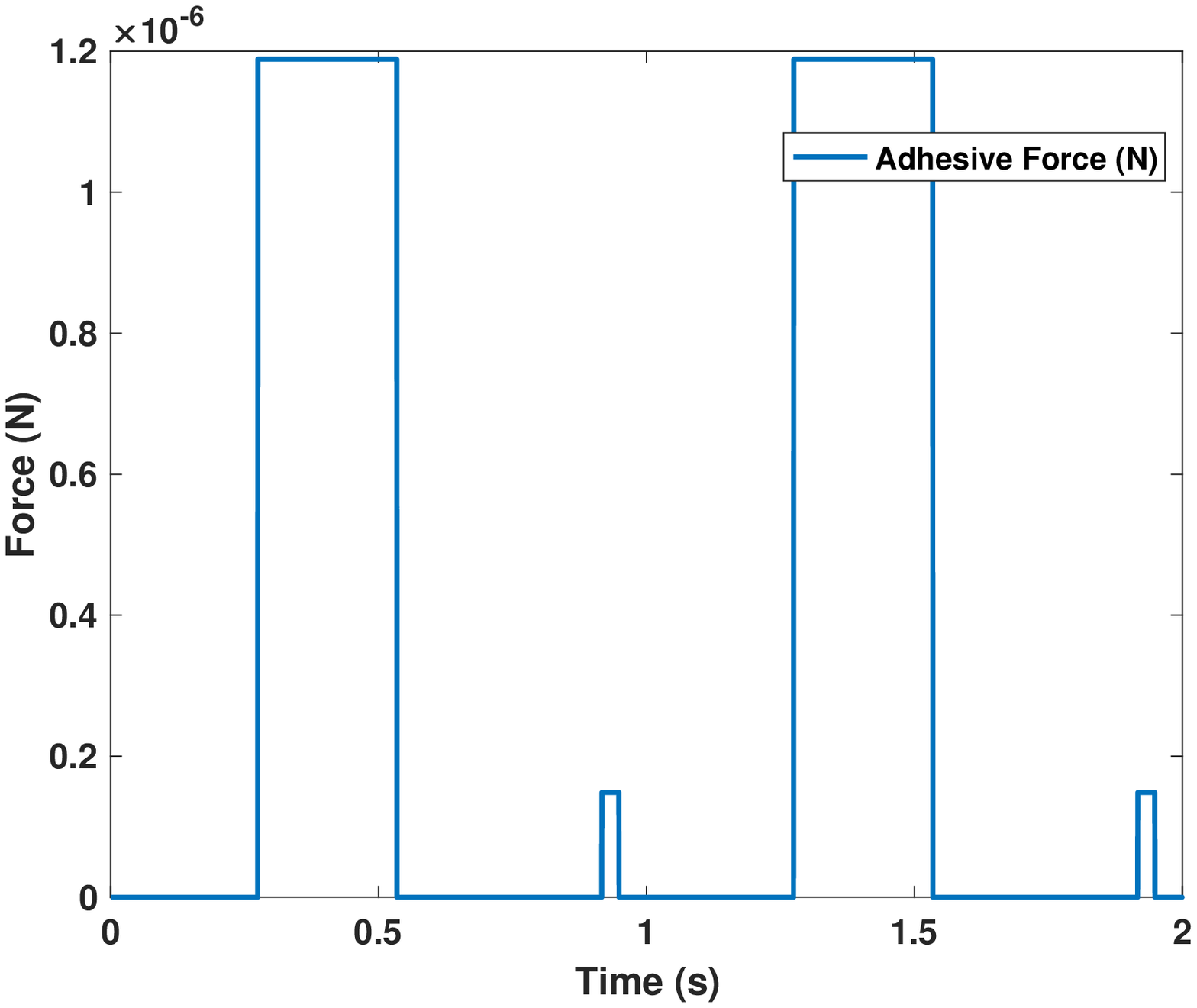}
\caption{Simulation result for adhesive force acting on $\mu$TUM robot when it is tumbling over the incline (paper) of 45$^{\circ}$ ($20mT$ field at $1Hz$). }
\label{figure:ADH} 
\end{figure} 
We applied a rotating magnetic field of $10 mT$ to the robot. The initial configuration the robot is ${\bf q} = [0,0,q_z,1,0,0,0]^{T}$, where the z-axis height of the CM $q_z = 100 \mu m$. The initial generalized velocity ${\bf V}$ is zero. As shown in Table~\ref{table_1}, the magnitude of mass ($m$) and volume ($V_m$) is small, which is in the order of 1e-8kg and 1e-11 $m^3$. To increase the accuracy in our simulations, we set the tolerance to be 1e-10. Furthermore, we scale up the metric units. In the simulations, the unit for mass is gram (g) and the unit for distance is millimeter (mm). 

Figure~\ref{figure:Tumbling} compares the experimental and simulation results with the ideal no-slip solution (Equation~\eqref{model_non_slip}). The simulation results match reasonably well with the experimental results up to the frequency of $10$ Hz. Furthermore, the simulation results also matches with the ideal solution since there is no slip at the contact during the tumbling motion (the slip velocity plotted in Figure~\ref{figure:Tumbling}, is almost zero and is visually indistinguishable from the $x$-axis). 
At the frequency of $15$ Hz, the experimentally obtained average velocity is higher than expected due to complications in the MagnebotiX machine producing the external magnetic field. It is suspected that stray field gradients become more prominent at higher rotational frequencies and pull the microrobot towards the edges of the workspace, causing it to move faster.

\subsection{Inclined Plane Traversal Tests on Paper}

\begin{table}[!tp]
\centering
\caption{Inclined plane tests on paper ($20mT$ $@$ $1Hz$).}
\begin{tabular}{c c c}
\hline \hline
Incline ($\varphi$)   & Simulation (Y/N) & Experiment (Y/N) \\
\hline
$5^{\circ}$ & Y &   Y    \\
$10^{\circ}$ & Y &   Y    \\
$15^{\circ}$ & Y &   Y    \\
$30^{\circ}$ & Y &   Y    \\
$45^{\circ}$ & Y &   Y    \\
$60^{\circ}$ & N &   N    \\
\hline
\end{tabular}
\label{table_paper}
\end{table}

In the second scenario, the simulation is used to determine whether the designed microrobot can climb an inclined surface (paper in dry conditions) at various angles. We applied a $20 mT$ rotating magnetic field and $1Hz$ frequency to the robot. We compared the simulation results with experimental results to validate our model. In this scenario, we again take the adhesive force into account in the simulation. The results are reported in Table~\ref{table_paper}. Based on the experimental result, the robots can go over a maximum inclination of $45^{\circ}$ on paper but it will fail to climb a slope of $60^{\circ}$. The simulation output matches these results. Figure~\ref{figure:ADH} plots the adhesive force when the robot is tumbling over the incline at $45^{\circ}$. It can be observed from this figure that the force changes periodically, depending on the contact surface. When the contact area is large (Length $\times$ Width), the adhesive force reaches a value of $3.8e-7 N$. When the contact area is small (Width $\times$ Height), the adhesive force value goes to $4.7e-8N$. In line contact cases, the adhesive force is almost zero.

\subsection{Inclined Plane Traversal Tests on Aluminum}
\begin{table}[!tp]
\caption{Parameters for improved $\mu$TUM on aluminum.}
\begin{tabular}{c c c}
\hline \hline
Description   & Value & Units \\
\hline
Mass (m)  & $4.44\times 10^{-8}$  & kg \\
Electrostatic Force ($F_{elect}$) & $0$  & N\\
Friction Coefficient ($\mu$) & $0.54$  & -\\
Magnetic Alignment Offset ($\alpha$) & $0$  & degree\\
Magnetic Volume  ($V_m$) & $3.2\times 10^{-11}$  & $m^3$\\
Magnetization ($|{\bf E}|$) & $51835$  & $A/m$\\
Coefficient of adhesion force ($C$) & 26.18 &$N/m^2$ \\
\hline
\end{tabular}
\label{table_3}
\end{table}

\begin{table}[!tp]
\centering
\caption{Inclined plane tests on aluminum ($20mT$ $@$ 1Hz).}
\begin{tabular}{c c c}
\hline \hline
Incline ($\varphi$)   & Simulation (Y/N) & Experiment (Y/N) \\
\hline
$30^{\circ}$ & Y &   Y    \\
$45^{\circ}$ & N &   N    \\
\hline
\end{tabular}
\label{table_4}
\end{table}

In our third scenario, we analyze the performance of a $\mu$TUM with improved magnetic properties. In Table~\ref{table_3}, the magnetization of the newer $\mu$TUM's ($ 51,835 \  A/m$) is much higher than that of original $\mu$TUM ($ 15,000 \  A/m$). Furthermore, the newer $\mu$TUM has zero magnetic alignment offset angle. The inclined tests are performed on aluminum, which is non-magnetic and conductive. Therefore, there shouldn't be any significant electrostatic force or additional magnetic force acting on the robot when it is tumbling over the substrate. Although an electromagnetic drag force may be exerted on the $\mu$TUM due to eddy currents induced in the conductive aluminum, this force is estimated to be two orders of magnitude smaller than the magnetic torque and thus negligible. The coefficient of adhesive force on aluminum was found to be $26.18 N/m^2$ and the coefficient of friction was found to be $0.54$.   These values are almost seven times more than the case for paper.  Therefore, we must also consider this adhesive force in our simulation. The procedure for obtaining the parameters is stated in our experimental setup section. 
 We applied a $20mT$ rotating magnetic field at $1Hz$ frequency to the robot. The result of inclined plane climbing tests are reported in Table~\ref{table_4}. In both the simulations and the experiments, the robot can successfully climb the inclination of $30^{\circ}$ but fails to climb it at $45^{\circ}$.



\section{Dynamic Simulation-Based Microrobot Design}

\begin{figure*}[!thbp]%
\centering
\begin{subfigure}[b]{0.66\columnwidth}
\includegraphics[width=\columnwidth]{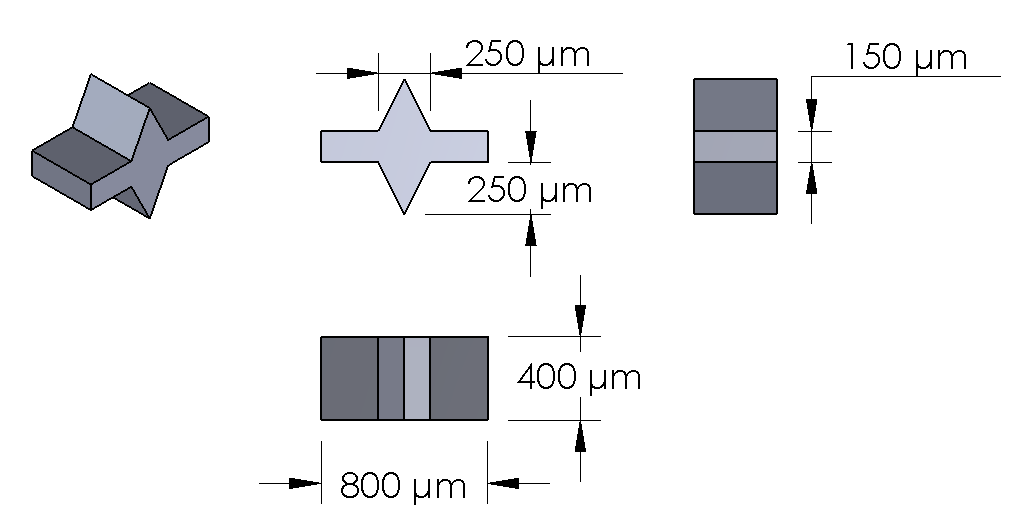}%
\caption{Spiked-shaped (SS) robot.}
\label{figure_spiked} 
\end{subfigure}\hfill%
\begin{subfigure}[b]{0.66\columnwidth}
\includegraphics[width=\columnwidth]{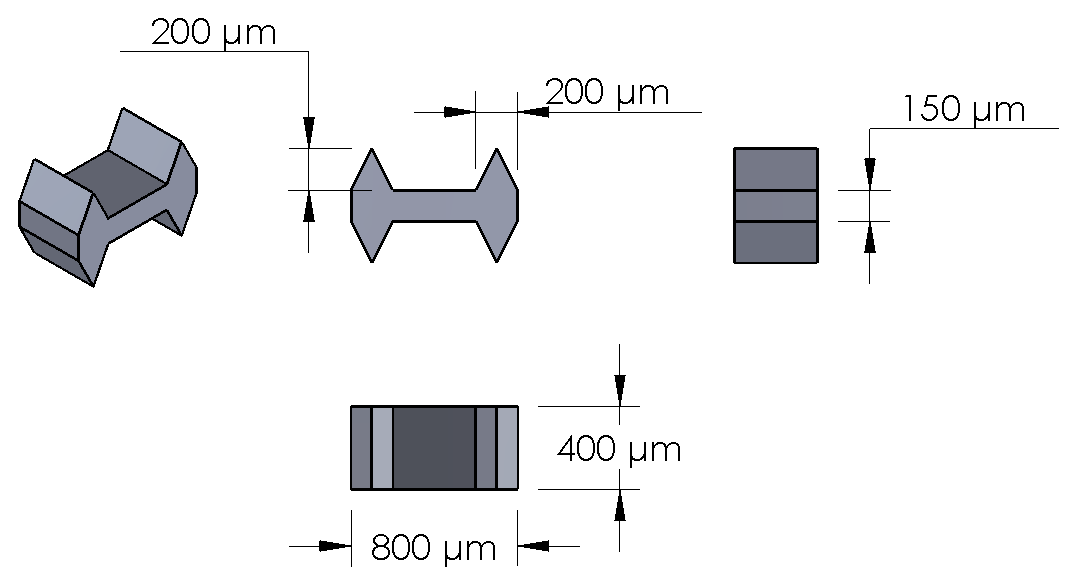}%
\caption{Spiked ends-shaped (SES)  robot. }
\label{figure_spiked_ends} 
\end{subfigure}\hfill%
\begin{subfigure}[b]{0.66\columnwidth}
\includegraphics[width=\columnwidth]{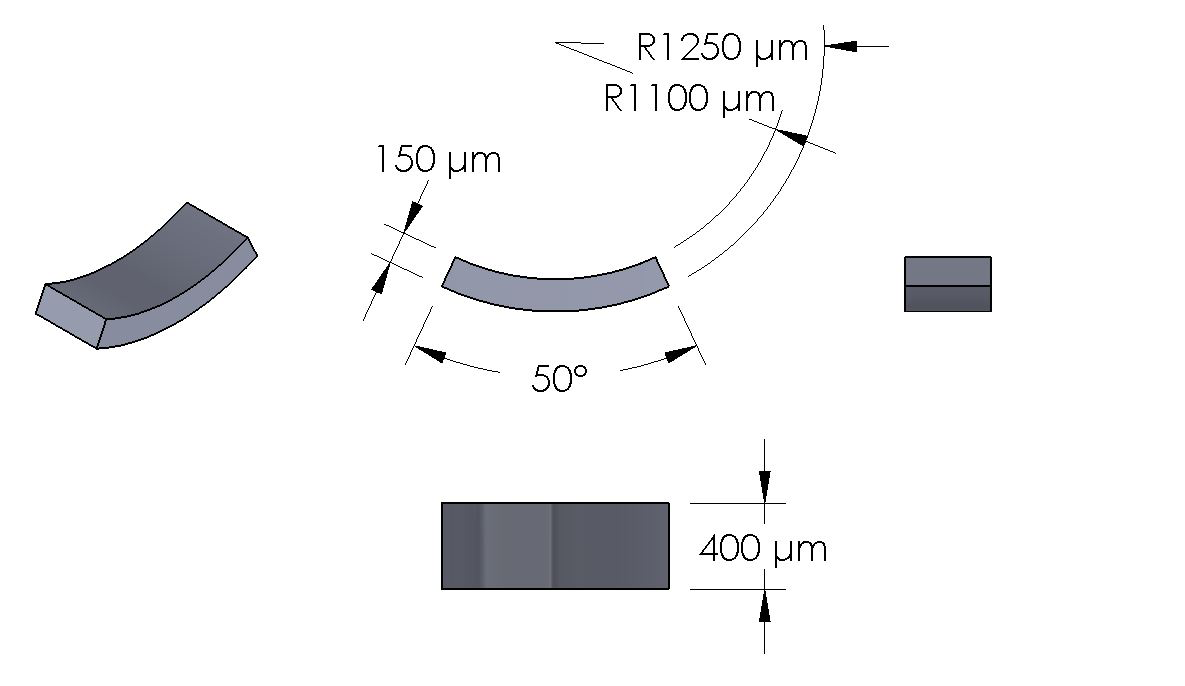}%
\caption{Curved-shaped robot. }
\label{figure_curved} 
\end{subfigure}\hfill%
\caption{Design and dimensions of $\mu$TUM robots with different geometric shapes. }
\label{Dimensions of robots}
\end{figure*}

Now that the simulation model has been validated, it can be used to explore alternative $\mu$TUM geometries for improved mobility. The priority for these new geometries is reducing or eliminating area contact during the tumbling cycle and decreasing the minimum actuation torque necessary to counteract adhesion force. This change can be implemented by altering the side profile of the microrobot and including spiked or curved features to prop the microrobot above the substrate surface. As shown in Figure~\ref{Dimensions of robots}, we simulated (a) spiked-shaped robots (SS), (b) robots with spiked ends (SES), (c) curved-shaped robots, and (d) the cuboid-shaped $\mu$TUM robots from before. To explore the effect of the robots' design and dimensions on their performance, we assume all the robots have the same geometry-independent properties listed in Table~\ref{table_3}, such as friction coefficient, magnetization, and coefficient of adhesion force. The geometry-dependent properties such as mass, magnetic volume, and moment of inertia change between designs.
The simulation includes both the tumbling locomotion tests and the inclined plane traversal tests from before. For each robots' design, those tests are performed once in the simulation. 
Since the microrobot moves on the paper or on the aluminum, we consider the adhesive forces in our simulation. 

In the tumbling locomotion tests, we applied a 20 $mT$ rotational magnetic field at 10 $Hz$ frequency to all the robots. Although these tests could have been performed at 1 $Hz$ for consistency, we increased this value to 10 $Hz$ in order to emphasize the velocity differences between the four designs due to slip. Initially, all robots stay at rest on the substrate surface. The initial configuration of each robot is ${\bf q} = [0,0,q_z,1,0,0,0]^{T}$, where the z-axis height of the CM, $q_z = 325 \mu m$ for SS,  $q_z = 275 \mu m$ for SES, $q_z = 220 \mu m$ for curved shape and $q_z = 100 \mu m$ for cuboid shape. The initial generalized velocity ${\bf V}$ is zero. 

\begin{figure}[!tbp]%
\centering
\begin{subfigure}[b]{0.8\columnwidth}
\includegraphics[width=\columnwidth]{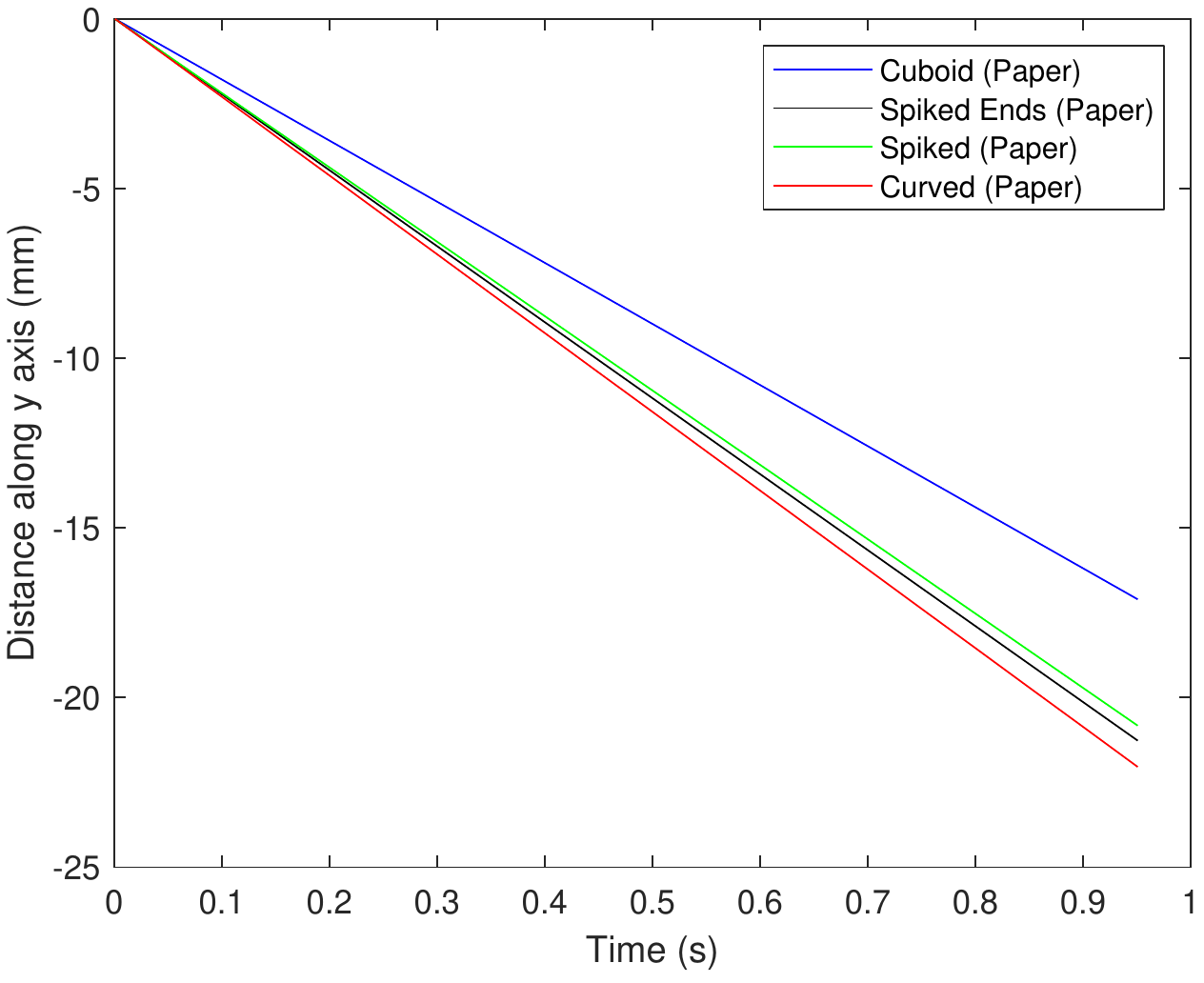}%
\label{paper_mul} 
\end{subfigure}
\caption{Simulation result for tumbling locomotion tests ($20mT$ field at $10Hz$) for robots with different geometric shapes on paper. The result for the tests on  aluminum are similar to the results on paper.}
\label{figure:TUMB_MUL}
\end{figure}

Figure~\ref{figure:TUMB_MUL} shows the displacement $q_y$ of all the robots along the $y$-direction as the robots tumble forward on a paper substrate. 
We only present the plots for paper, since, each robot's performance on paper is similar to it on aluminum. Furthermore, the curved shape robot was found to move the fastest while the cuboid shape robot moved the slowest.
 
 In inclined plane traversal test, we chose the substrate to be aluminum. The initial configuration ${\bf q} = [0,0,q_z/\cos(\varphi),\cos(\varphi/2),\sin(\varphi/2),0,0]^{T}$, where the inclined angle $\varphi=20^{\circ}, 30^{\circ}, 45^{\circ}$. All robots except the curved shape robot successfully climbed the incline up to $30^{\circ}$ and failed to climb it at $45^{\circ}$.  The result of inclined plane climbing tests are reported in Table~\ref{table_5}. Based on the simulation results, we can conclude that the curved shaped robot performs best in terms of linear speed, but is comparatively worse at climbing. In addition, we find that the basic cuboid shape robot is not the best design for tumbling locomotion. Instead, we found that robots with spiked ends geometry (SES) has the best overall performance in locomotion tests and inclined plane tests.



\begin{table}[!tp]
\centering
\caption{Simulation results for robots with different geometric shapes: inclined plane tests on aluminum ($20mT$ $@$ $1Hz$).}
\begin{tabular}{c c c c c}
\hline \hline
Incline ($\varphi$)   & Cuboid  & Spiked  & Spiked Ends  & Curved  \\
 & (Y/N) & (Y/N)& (Y/N)&(Y/N)\\
\hline
$20^{\circ}$ & Y &   Y  &Y  &Y \\
$30^{\circ}$ & Y &   Y  &Y  &N\\
$45^{\circ}$ & N &   N  &N  &N\\
\hline
\end{tabular}
\label{table_5}
\end{table}

\comment{\hfill%
\begin{subfigure}[b]{0.8\columnwidth}
\includegraphics[width=\columnwidth]{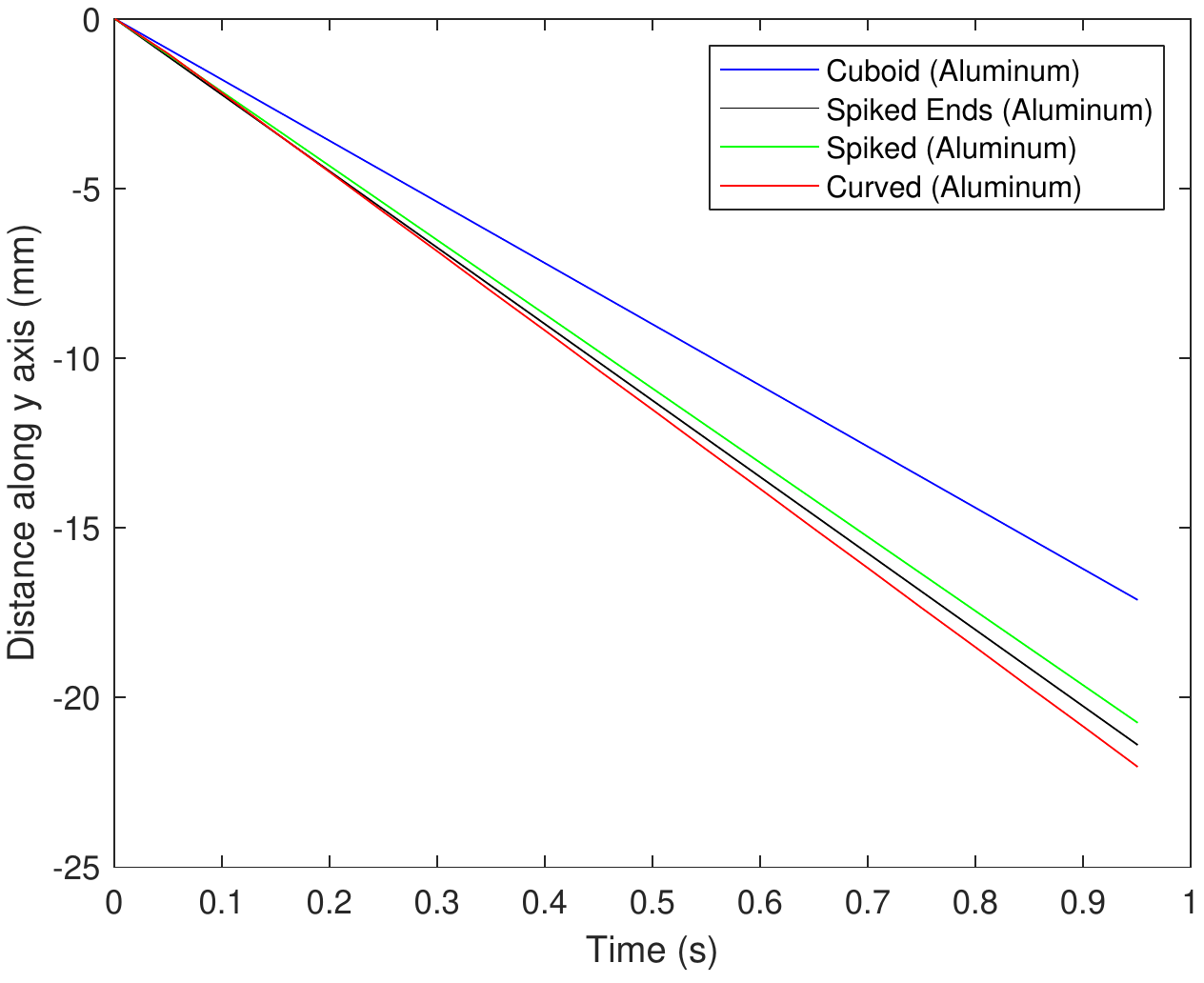}%
\label{aluminum_mul} 
\end{subfigure}\hfill%
}
\section{Manufacturing Alternative Microrobot Geometries}
\subsection{Manufacturing Methods \& Considerations}
\label{se:Manufacture_consideration}
To verify the results of the shape exploration process, we fabricated physical versions of the spike-shaped robots (SS) and spiked ends-shaped robots (SES) for further experimentation. Several manufacturing challenges and limitations, however, were encountered while producing these geometries.

While capable of fabricating precise designs within the nanometer range, photolithography is limited to patterning 2D designs over a flat substrate. Complex geometric features can only be patterned along one direction and surfaces along other directions maintain a rectangular cross-section. For the basic cuboid shape microrobot, the top plane (a 800$\mu$m $\times$ 400$\mu$m rectangle) is the face that is patterned during the photolithography process. To realize the more complex spike features, however, the side plane of the microrobot must be patterned instead. This requirement poses problems because the microrobot is much wider (400$\mu$m) than it is tall (100$\mu$m), necessitating a thicker SU-8 layer to pattern from. Spincoating a single, uniform SU-8 layer thicker than 200$\mu$m is a challenging and uncommon procedure, with multiple stacked SU-8 layers being the preferred alternative. However, this option is not available, for the magnetic microrobots. The embedded neodymium particles are opaque and block the vision of the preceding SU-8 layers underneath, making optical mask alignment difficult between separate SU-8 layers. An attempt was made to fabricate multiple cross-sectional slices of thin SU-8 layers and manually bind them together to form a completed microrobot. This process was prone to alignment errors between the slices due to its manual nature.  It also led to rough edges on the resultant microrobots that impaired consistent motion.
Additional difficulties were encountered while separating the patterned spiked SU-8 geometries from their silicon wafer substrates. The small feature size of the spikes, additional stress concentration points, and brittle nature of SU-8, made breakages common during microrobot extraction and handling operations.

To circumvent these manufacturing challenges, we proposed an alternative method of fabricating the desired microrobot geometry. Instead of rigid SU-8 photoresist, more compliant, elastomeric PDMS (Polydimethylsiloxane, Sylgard 184) is used to make the microrobot more robust against fracture. While PDMS is generally considered to be a soft material, we argue that the applied loads are too small to yield significant deformation of the microrobot and the simulation's rigid body assumption can be maintained. Unlike SU-8, PDMS cannot be patterned using photolithography and requires a mold or an alternative process to be shaped into the desired geometry. Since molds tend to produce burrs on the edges of the extracted geometry, which can result in erratic tumbling motion, we opted to use laser cutting to form the PDMS material instead.

Making laser cutting a viable fabrication option required additional alterations to the microrobot design to ensure clean cuts. The PDMS material can undergo excessive curing and become extremely brittle if the embedded neodymium particles absorb too much thermal energy and significantly raise internal temperature. To avoid this problem, microrobots were patterned at half their intended size and the mass ratio of embedded magnetic particles was reduced from 1:1 to $\approx$1:8.5. This particle reduction also lead to a proportional decrease in microrobot magnetization from $ 51,835 \  A/m$ to $ 18,661 \  A/m$.

The fabrication process for the resultant spiked geometry microrobots (SES and SS) consists of the following steps: The PDMS is thoroughly mixed with the magnetic particles to remove air bubbles large enough to cause problems in the fabrication process. Then, the doped PDMS is placed on a glass microscope slide with a \#2 glass cover slip (Fisher Scientific) on each side to set the layer thickness (400$\mu$m), and another microscope slide on top to keep the layer uniform at the desired size. The microscope slides are held together using a binder clip on each side and the entire system is placed on a hot plate at 90\textdegree C for an hour and a half to cure the PDMS. Lastly, the thin film is removed from the glass, cut in the shape of the microrobots using a laser system, and then magnetized using the same process as the SU-8 microrobots (with a constant 9T magnetic field). For the cutting procedure, we utilize a custom laser cutter system consisting of a femtosecond laser (CARBIDE, 04-1000), beam expander (Thorlabs, BE02-050B), attenuator (Altechna, Watt Pilot), a waveplate, Brewster type polarizer, and a 20X objective lens (Mitutoyo, 0.42NA) ~\cite{akin2020dual}.

\subsection{Manufacturing Errors and Limitations}
\begin{figure}
\centering
\includegraphics[width=0.7\columnwidth]{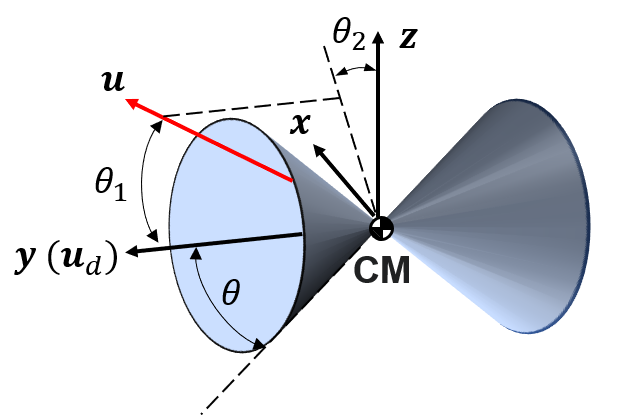}%
\caption{Magnetization error: The possible deviations of the magnetic axis from the ideal is approximated by a double cone with aperture $\theta = 10^{\circ}$. The axis of the cone is same as the $\bf y$ axis of the robot, and the center of the cone is located at the robot's CM. The ideal magnetic axis, ${\bf u}_d$, coincides with the $\bf y$ axis. The actual magnetic axis, $\bf u$, which is characterized by $\theta_1$ and $\theta_2$, lies on or within the cone.
}
\label{figure_magnet}
\end{figure}

\begin{figure}[!tbp]%
\centering
\includegraphics[width=\columnwidth]{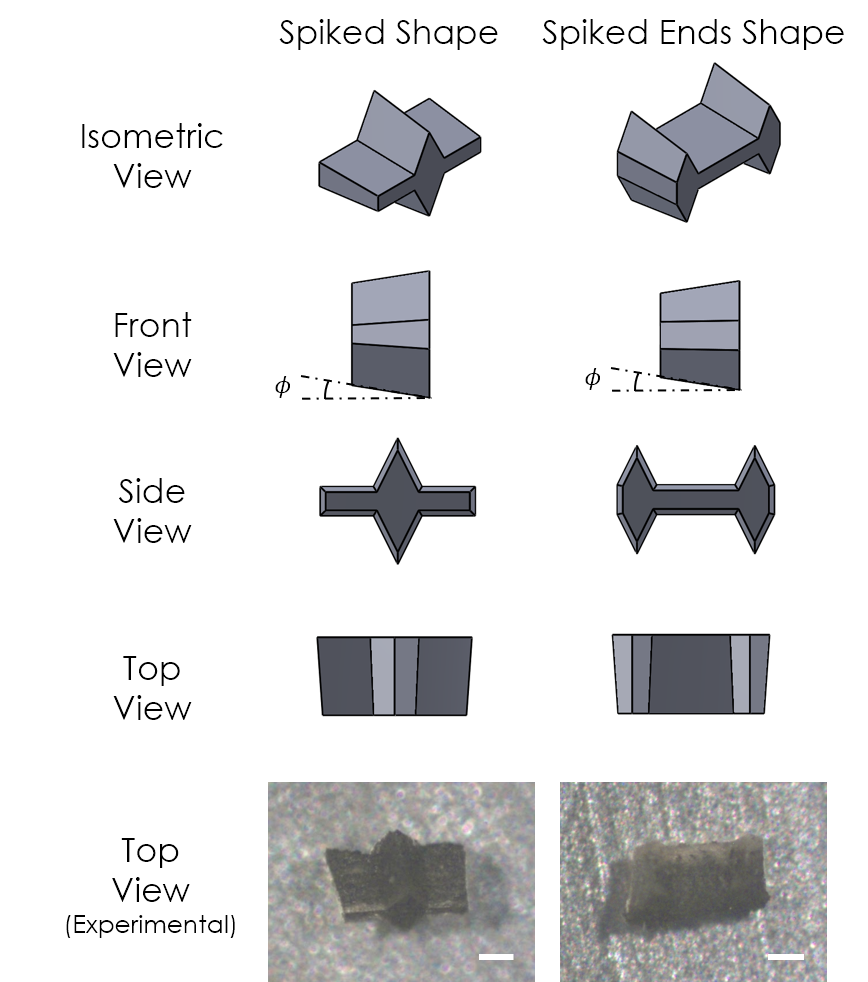}%
\caption{Geometric error: The imperfections in the geometry of the manufactured microrobots with spiked shapes and spiked ends. The geometry of the laser-cut tapered edges is approximated by using  an inward draft.}
\label{figure_draft_angle} 
\end{figure}

When fabricating the microrobots, there can be errors due to: (1) deviation of the magnetic axis from the ideal (we call this error {\em magnetic misalignment} or {\em magnetization error}, see Figure~\ref{figure_magnet}), and (2) imperfections in the geometry ({\em geometric error}) of the microrobots (see Figure~\ref{figure_draft_angle}). The laser beam used in the cutting process has tapered edges that result in an inward draft angle on the geometry of the fabricated microrobots. Additionally, the spiked features of the microrobots make them more difficult to manually align and mount during the magnetization process, leading to potential alignment errors in magnetic polarization. These manufacturing errors may affect the motion of robots, and thus understanding the effects of the errors is important in designing the robots. The change in material from SU-8 to PDMS can also affect interactions between the microrobot and substrate. The friction and adhesion coefficients of PDMS against aluminum, for example, are both substantially higher than those for SU-8.
The presence of these discrepancies can lead to unexpected tumbling trajectories and behaviors that were not predicted using the original simulation parameters.

\section{Incorporating Manufacturing Errors into Dynamic Model}

In this section, we perform simulation and experimental studies to understand the effects of the manufacturing errors.

\begin{figure*}[!htbp]%
\centering
\begin{subfigure}[b]{0.66\columnwidth}
\includegraphics[width=\columnwidth]{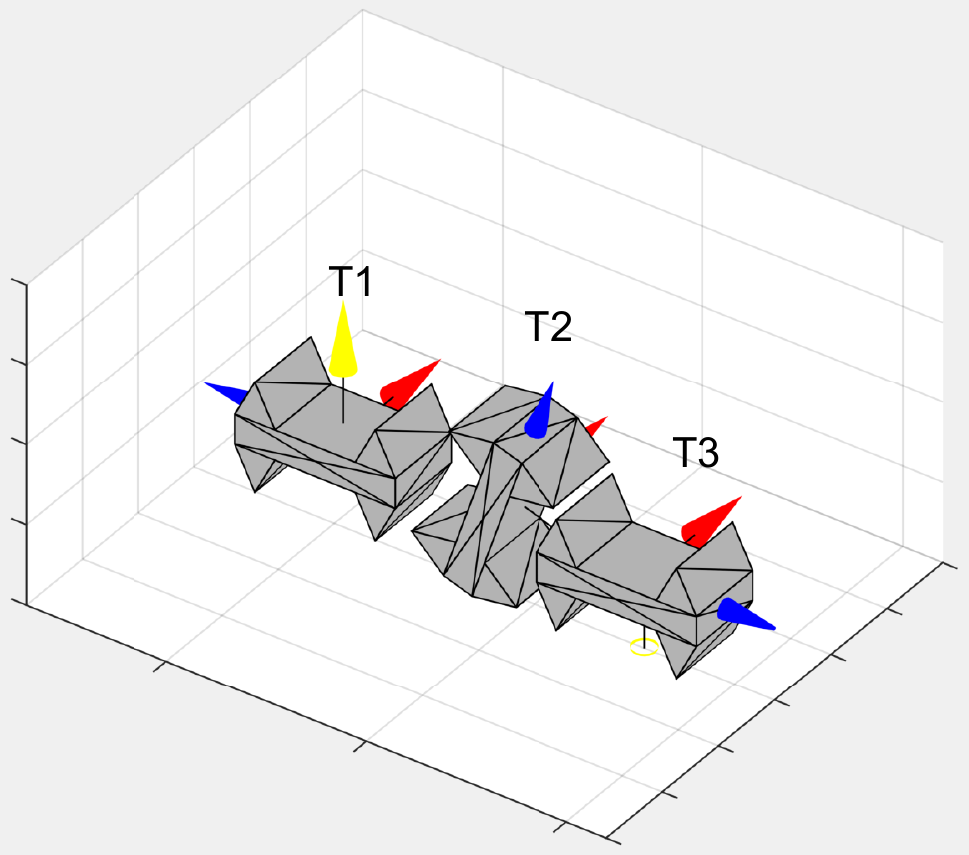}%
\caption{}
\label{SES_tumb} 
\end{subfigure}\hfill
\begin{subfigure}[b]{0.66\columnwidth}
\includegraphics[width=\columnwidth]{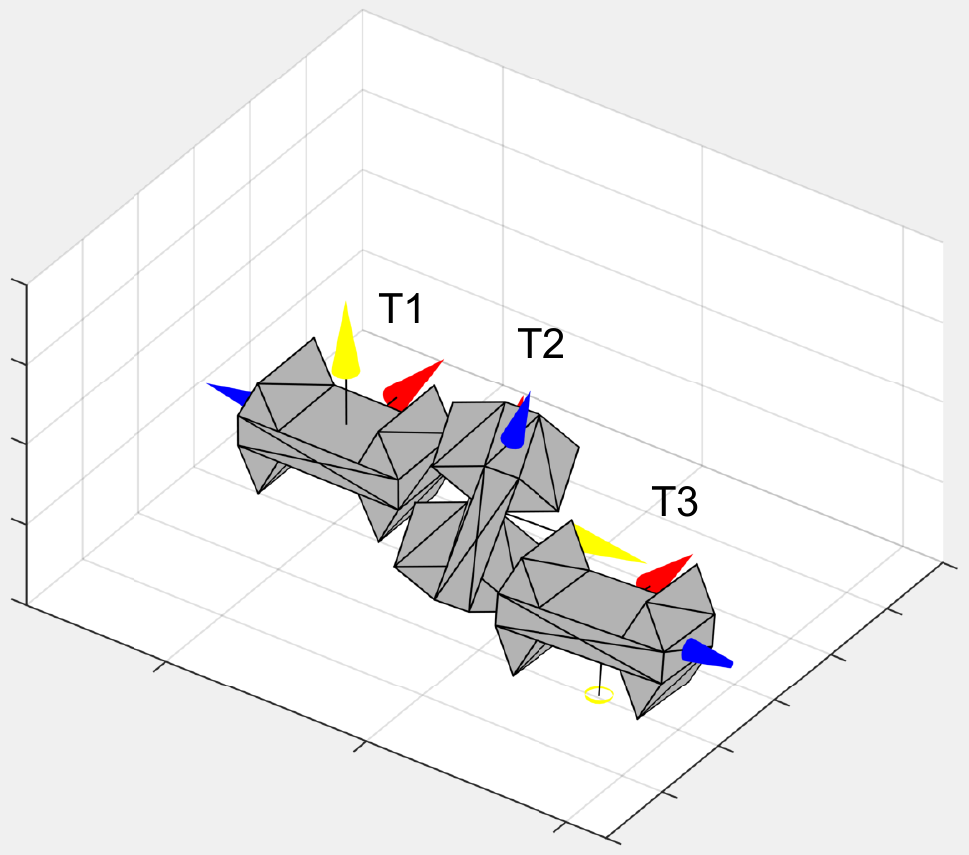}%
\caption{}
\label{SES_tilt} 
\end{subfigure}\hfill
\begin{subfigure}[b]{0.66\columnwidth}
\includegraphics[width=\columnwidth]{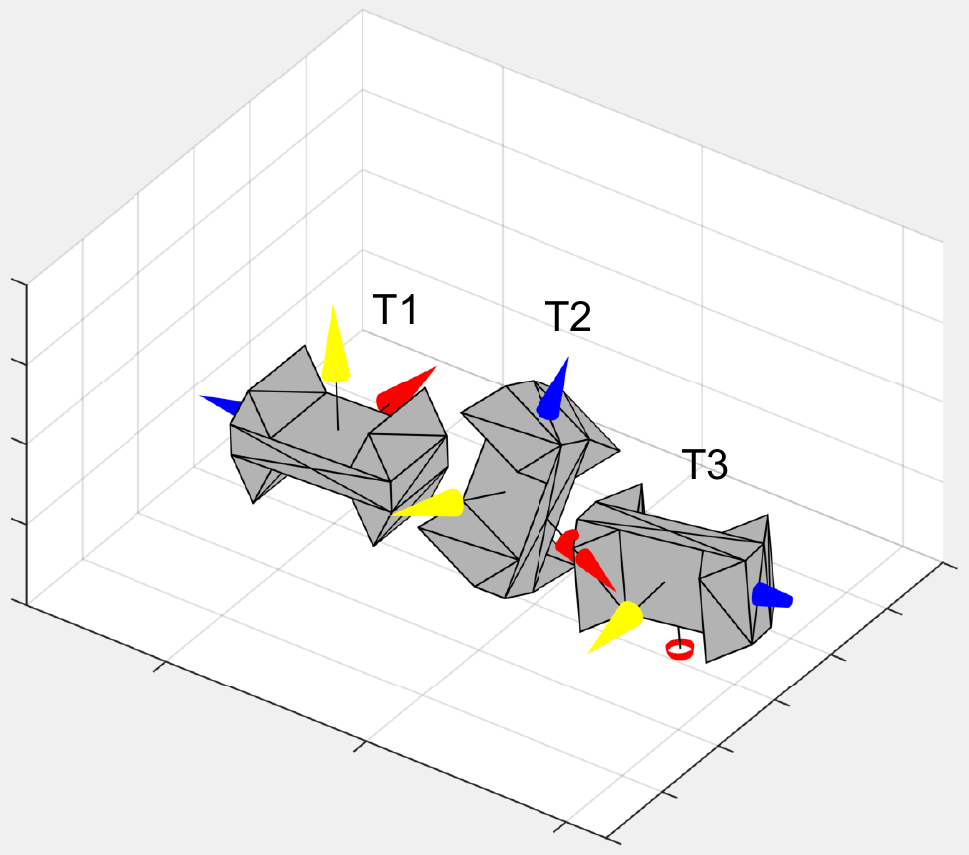}%
\caption{}
\label{SES_flip} 
\end{subfigure}
\caption{The tumbling motion of the SES microrobot when it (a) moves as expected, or (b) twists about the longitudinal axis of the body (the blue axis) with twist angle $-45^{\circ} \le \vartheta \le 45^{\circ}$ but falls on the spikes (as desired) (c) twists about the longitudinal axis of the body with  angle $\vartheta \ge 45^{\circ}$ or $\vartheta \le -45^{\circ}$ to fall on a flat face with no spikes. (Note: T1, T2, and T3 correspond to instances in time with T1 $<$ T2 $<$ T3.)}
\label{SES_tumbling}
\end{figure*}

\begin{figure*}[!htbp]%
\centering
\begin{subfigure}[b]{1\columnwidth}
\includegraphics[width=\columnwidth]{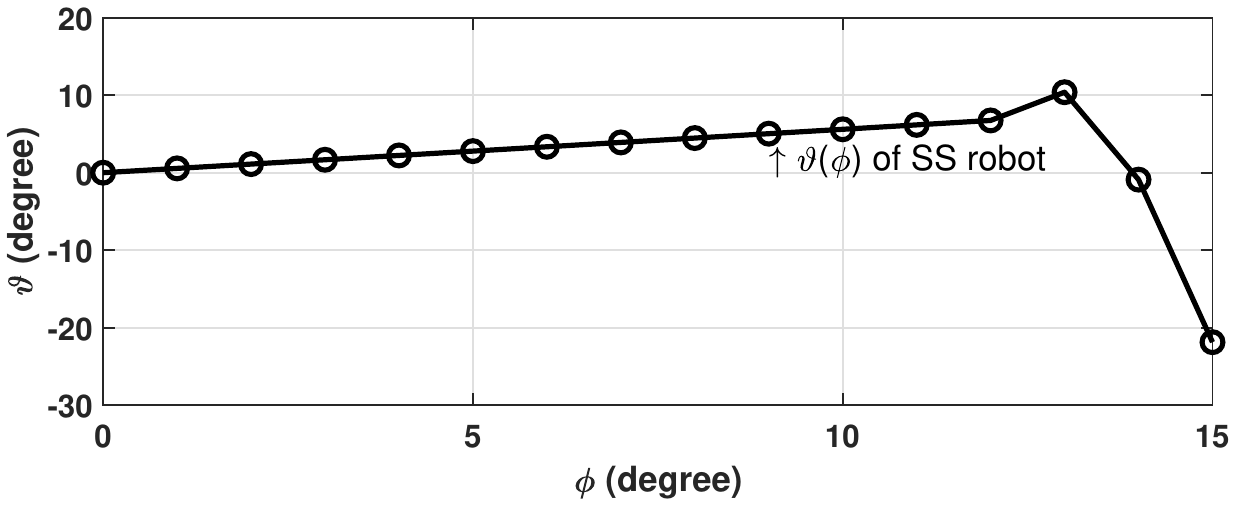}%
\caption{}
\label{SS_phi_pitch} 
\end{subfigure}\hfill
\begin{subfigure}[b]{1\columnwidth}
\includegraphics[width=\columnwidth]{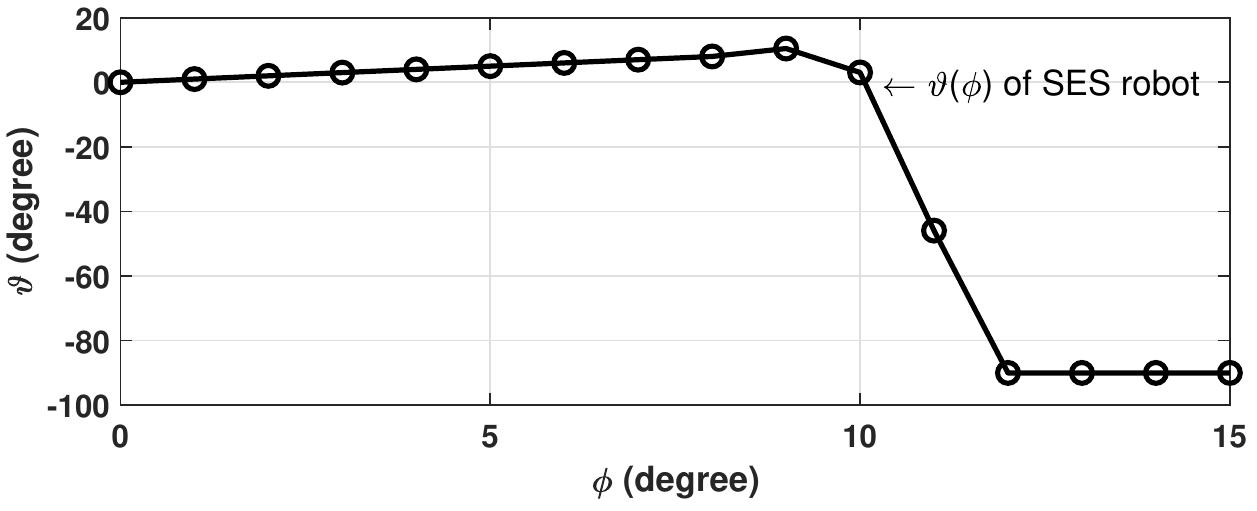}%
\caption{}
\label{SES_phi_pitch} 
\end{subfigure}\hfill
\begin{subfigure}[b]{1\columnwidth}
\includegraphics[width=\columnwidth]{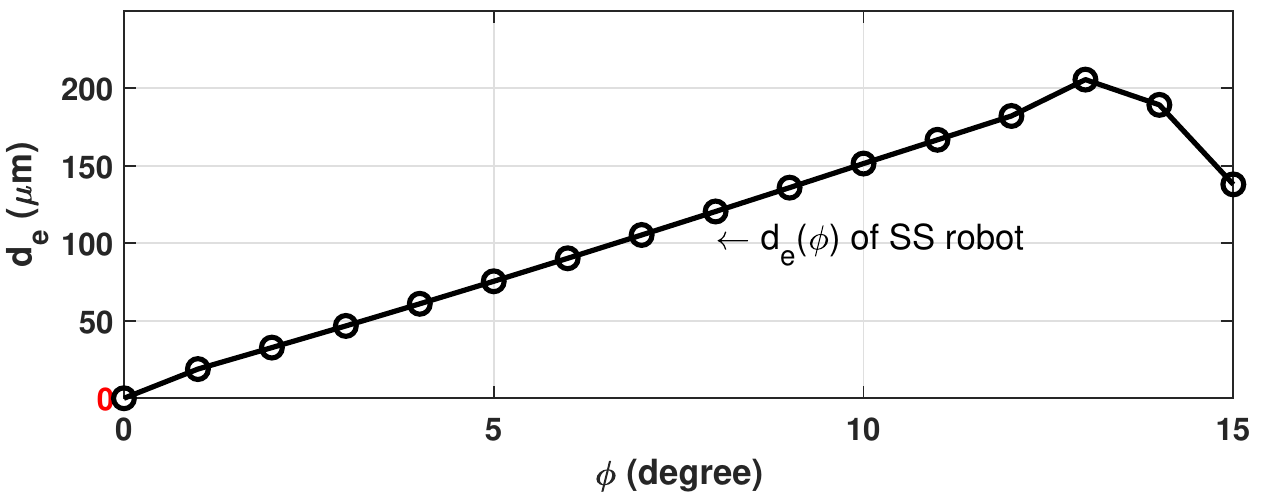}%
\caption{}
\label{SS_phi_d} 
\end{subfigure}\hfill
\begin{subfigure}[b]{1\columnwidth}
\includegraphics[width=\columnwidth]{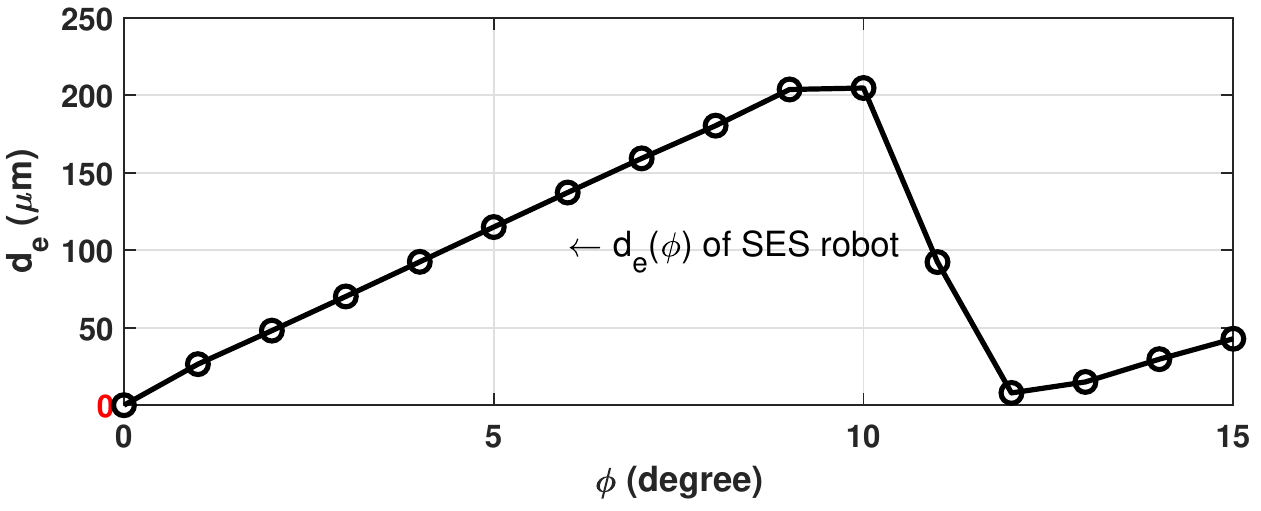}%
\caption{}
\label{SES_phi_d} 
\end{subfigure}\hfill
\begin{subfigure}[b]{1\columnwidth}
\includegraphics[width=\columnwidth]{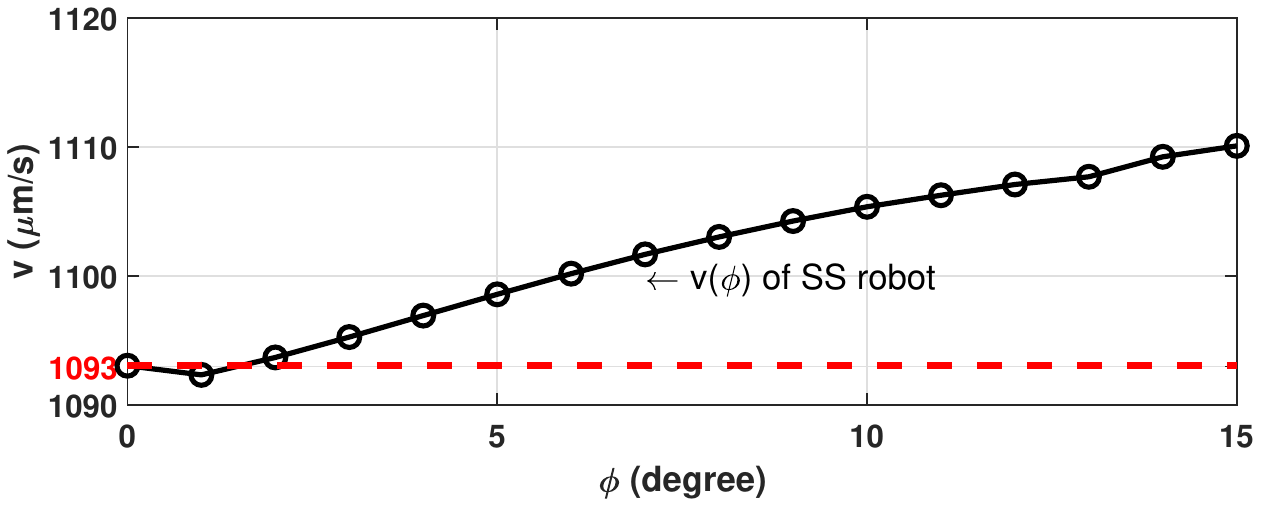}%
\caption{}
\label{SS_phi_v} 
\end{subfigure}\hfill
\begin{subfigure}[b]{1\columnwidth}
\includegraphics[width=\columnwidth]{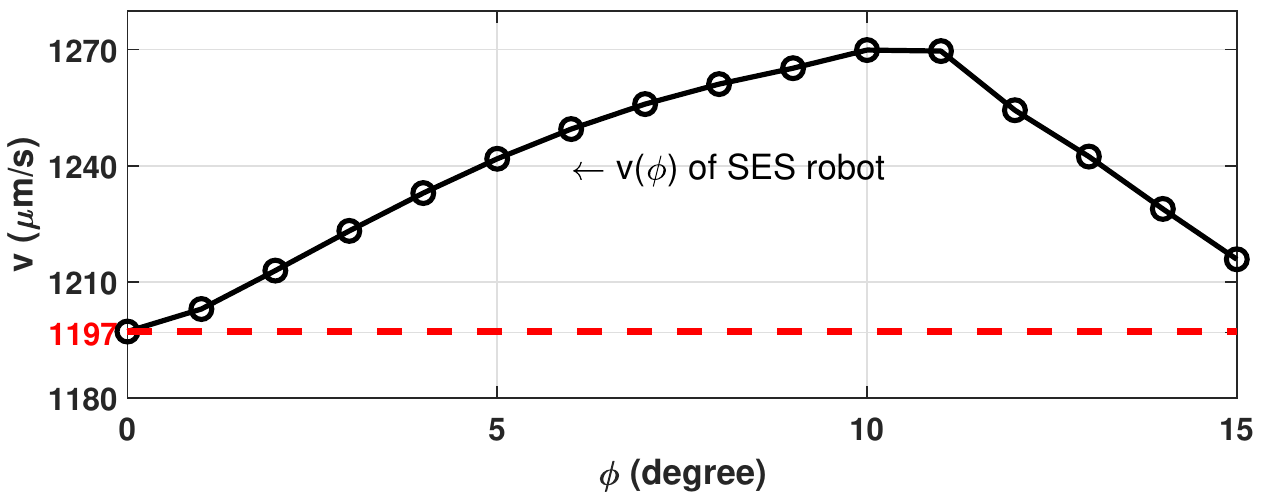}%
\caption{}
\label{SES_phi_v} 
\end{subfigure}\hfill
\caption{The simulation results for locomotion tests at 1 Hz with geometric errors of an inward draft angle $\phi = 0^{\circ},1^{\circ},...,15^{\circ}$. The simulation time is $1$s. In all $16$ simulation runs, the magnetization error is ignored ($\theta_1 = \theta_2 = 0^{\circ}$). The plots in the first column show the trends for SS microrobot of (a) angle of twist $\vartheta$ after one cycle of motion, (c) the drift $d_e$ after one cycle,  and  (e) the average translational speed $v$.  Similarly, the plots (b),(d) and (f) in second column shows the results for SES robots. The velocity values for the ideal situation, without manufacturing errors (shown in red dashed line), are v = $1093 \mu m/s$ and v = $1197 \mu m/s$, for the SS and SES designs, respectively. }
\label{analysis_phi_1_HZ} 
\end{figure*}


\begin{figure*}[!htbp]%
\centering
\begin{subfigure}[b]{1\columnwidth}
\includegraphics[width=\columnwidth]{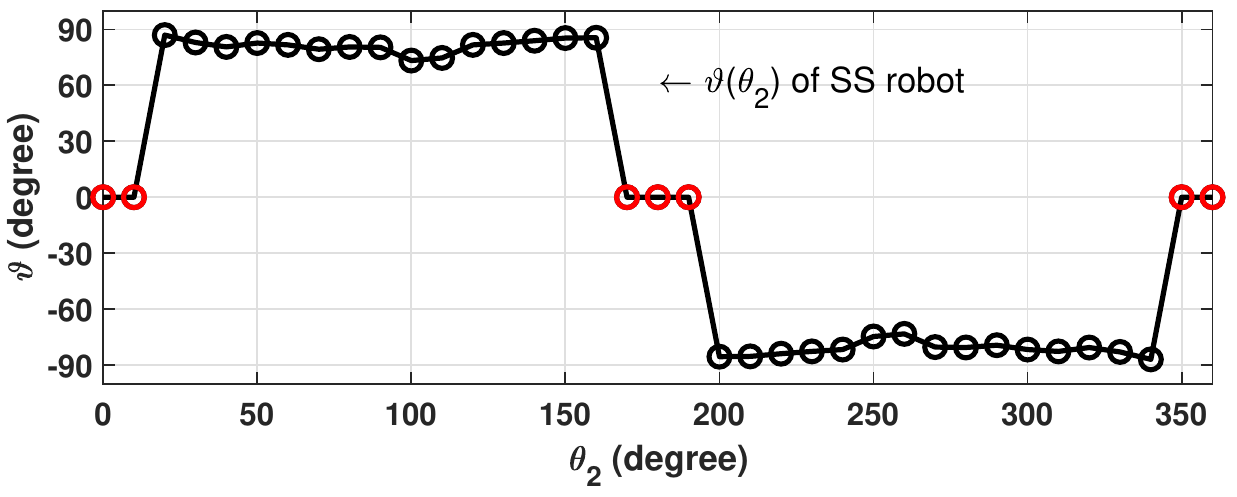}%
\caption{}
\label{SS_theta_pitch} 
\end{subfigure}\hfill
\begin{subfigure}[b]{1\columnwidth}
\includegraphics[width=\columnwidth]{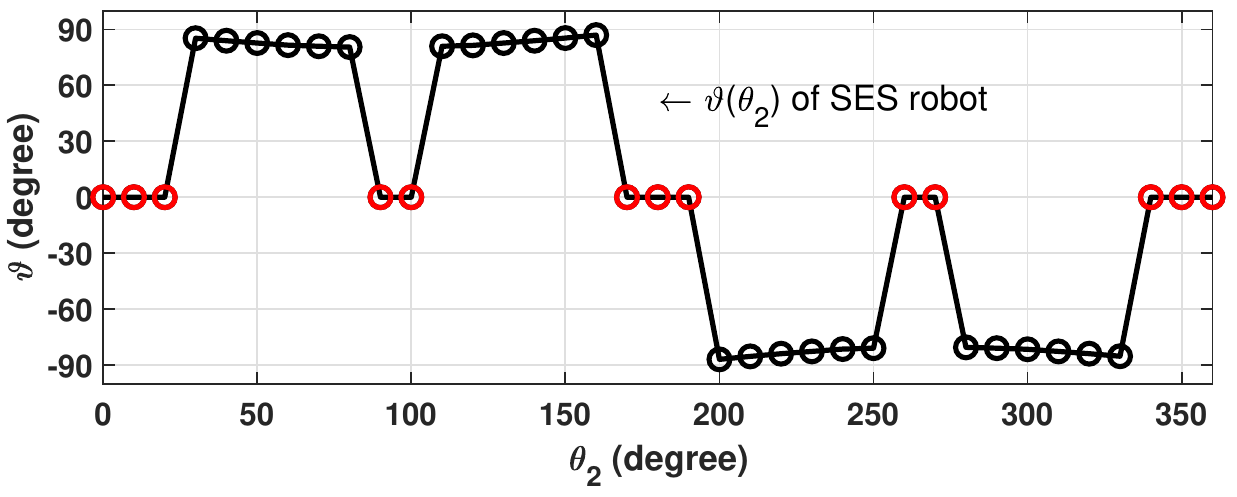}%
\caption{}
\label{SES_theta_pitch} 
\end{subfigure}\hfill
\begin{subfigure}[b]{1\columnwidth}
\includegraphics[width=\columnwidth]{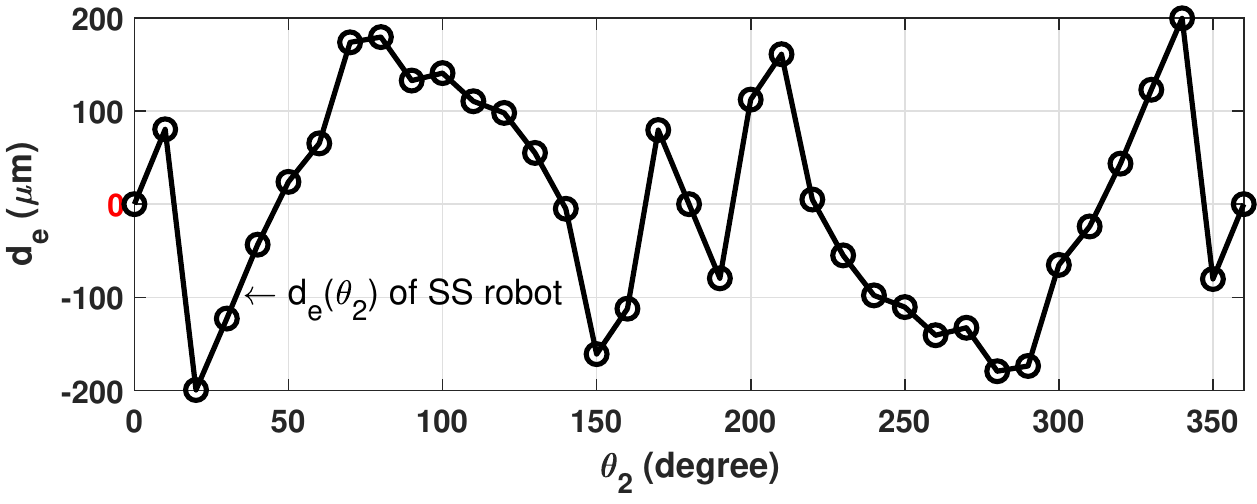}%
\caption{}
\label{SS_theta_d} 
\end{subfigure}\hfill
\begin{subfigure}[b]{1\columnwidth}
\includegraphics[width=\columnwidth]{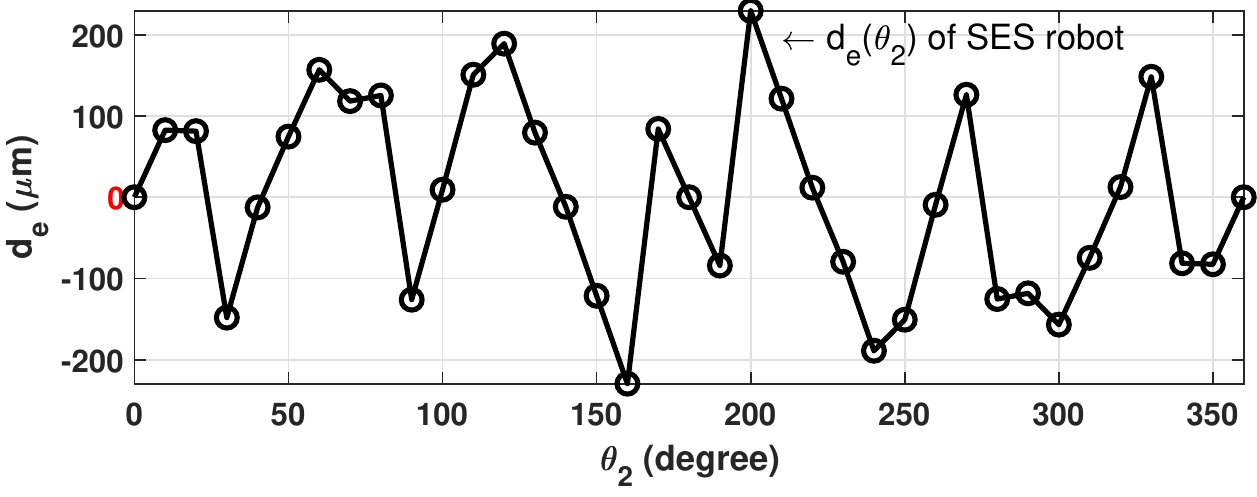}%
\caption{}
\label{SES_theta_d} 
\end{subfigure}\hfill
\begin{subfigure}[b]{1\columnwidth}
\includegraphics[width=\columnwidth]{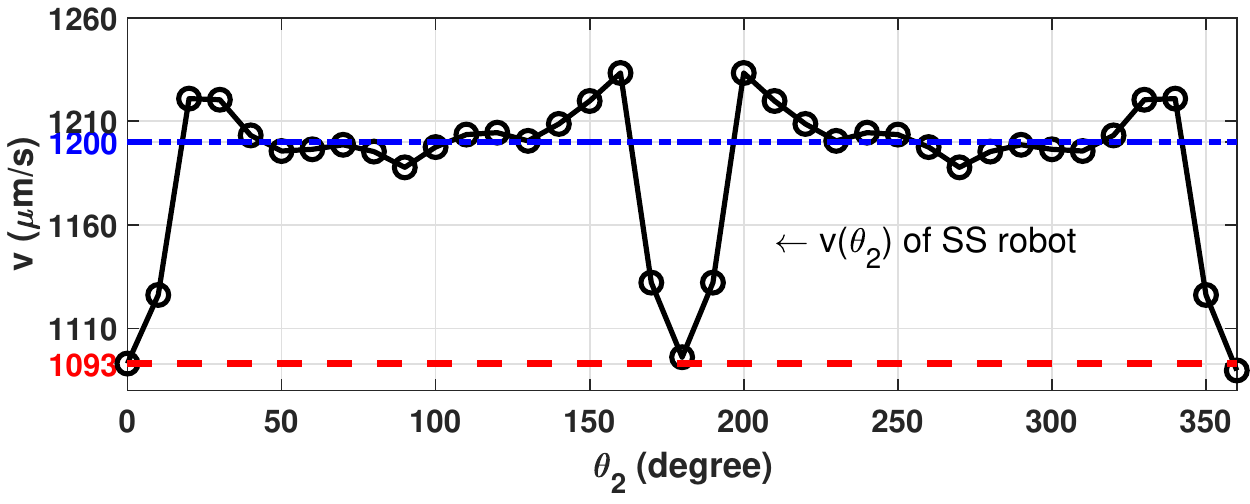}%
\caption{}
\label{SS_theta_v} 
\end{subfigure}\hfill
\begin{subfigure}[b]{1\columnwidth}
\includegraphics[width=\columnwidth]{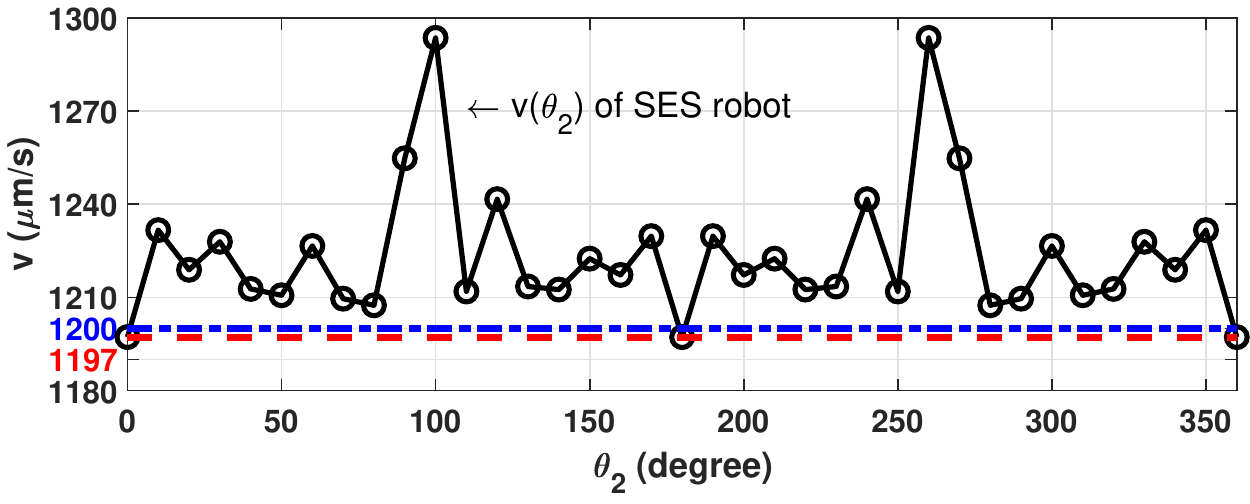}%
\caption{}
\label{SES_theta_v} 
\end{subfigure}\hfill
\caption{The simulation results for locomotion tests at $1$ Hz with magnetization error: $\theta_1 = 10^{\circ}$ and $\theta_2 = 10n^{\circ}, n=1,2,...,36$ (the simulation with $\theta_1 = 0^{\circ}$ and $\theta_2 = 0^{\circ}$ are also included). In all $37$ simulation runs, we ignore the imperfections in the geometry ($\phi = 0^{\circ}$). The plots in the first column show the trends for SS microrobot of (a) angle of twist $\vartheta$ after one cycle of motion, (c) the drift $d_e$ after one cycle,  and  (e) the average translational speed $v$.  Similarly, the plots (b),(d) and (f) in second column shows the results for SES robots. When the robot `flips', the value for the speed $v$ is near $1200 \mu m/s$ (shown in blue dashed line).}
\label{analysis_theta_1_HZ} 
\end{figure*}

\begin{figure*}
\begin{subfigure}[b]{1\columnwidth}
\includegraphics[width=\columnwidth]{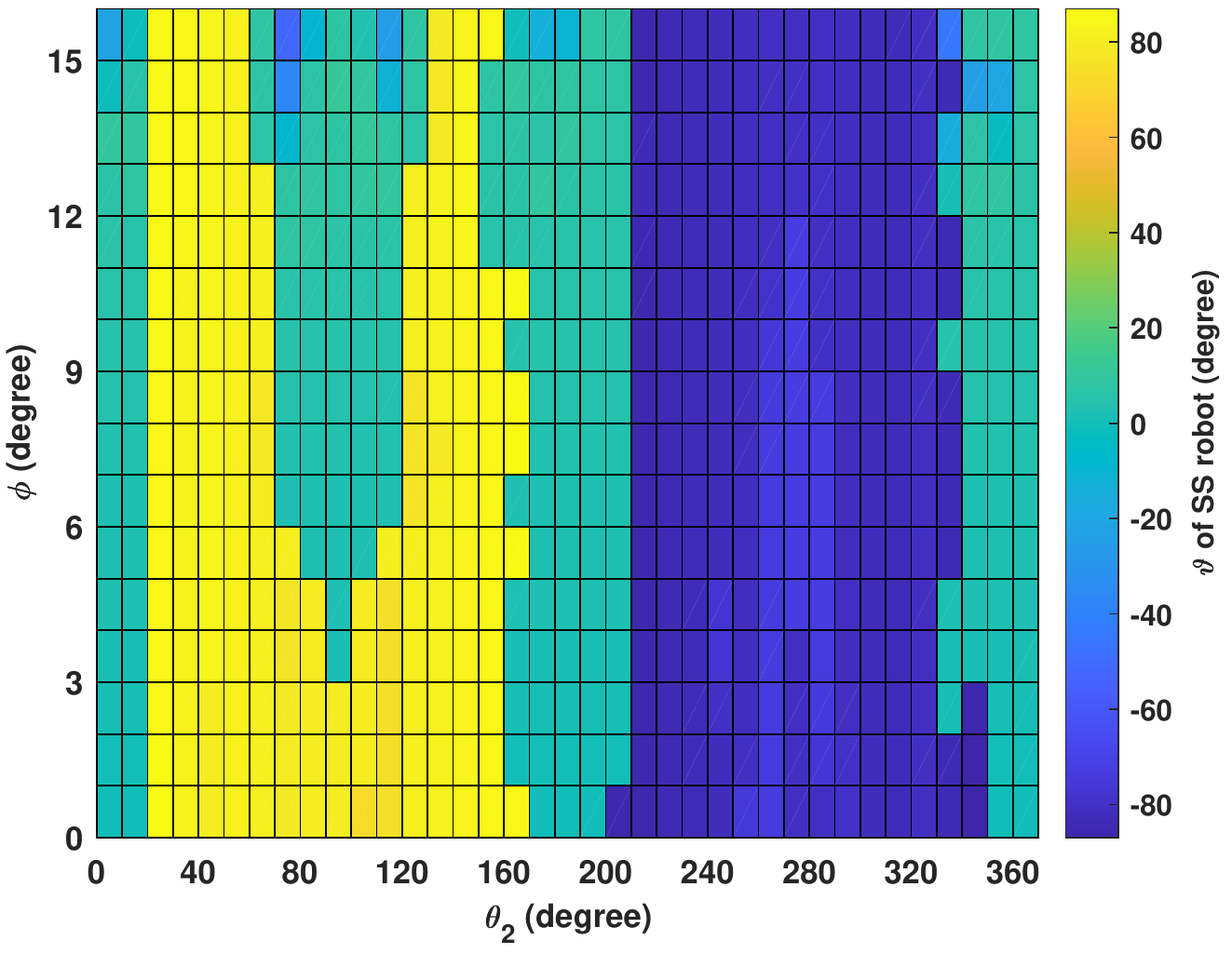}%
\caption{}
\label{analysis_pitch_ss_1_HZ} 
\end{subfigure}\hfill%
\begin{subfigure}[b]{1\columnwidth}
\includegraphics[width=\columnwidth]{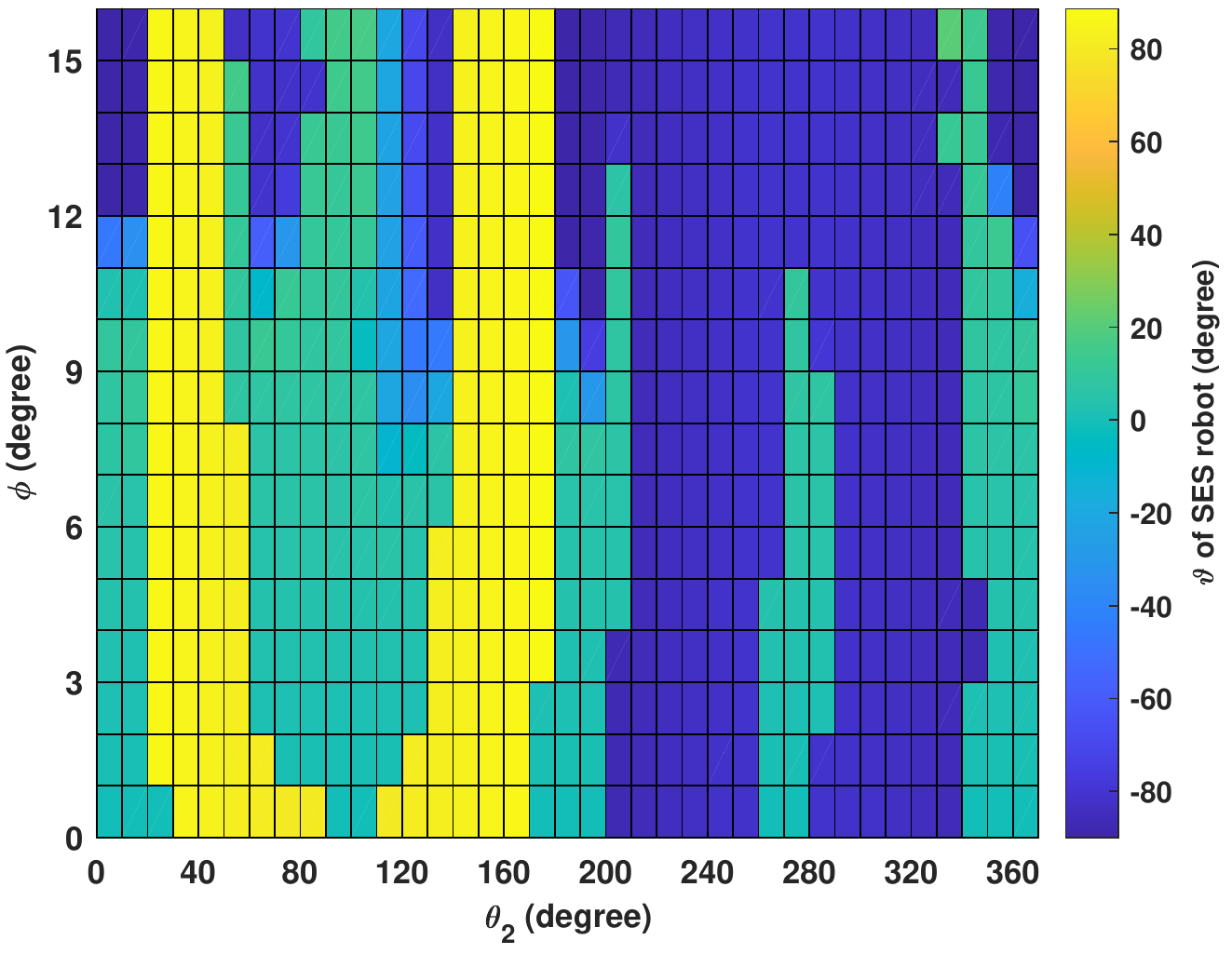}%
\caption{}
\label{analysis_pitch_ses_1_HZ} 
\end{subfigure}\hfill%
\begin{subfigure}[b]{1\columnwidth}
\includegraphics[width=\columnwidth]{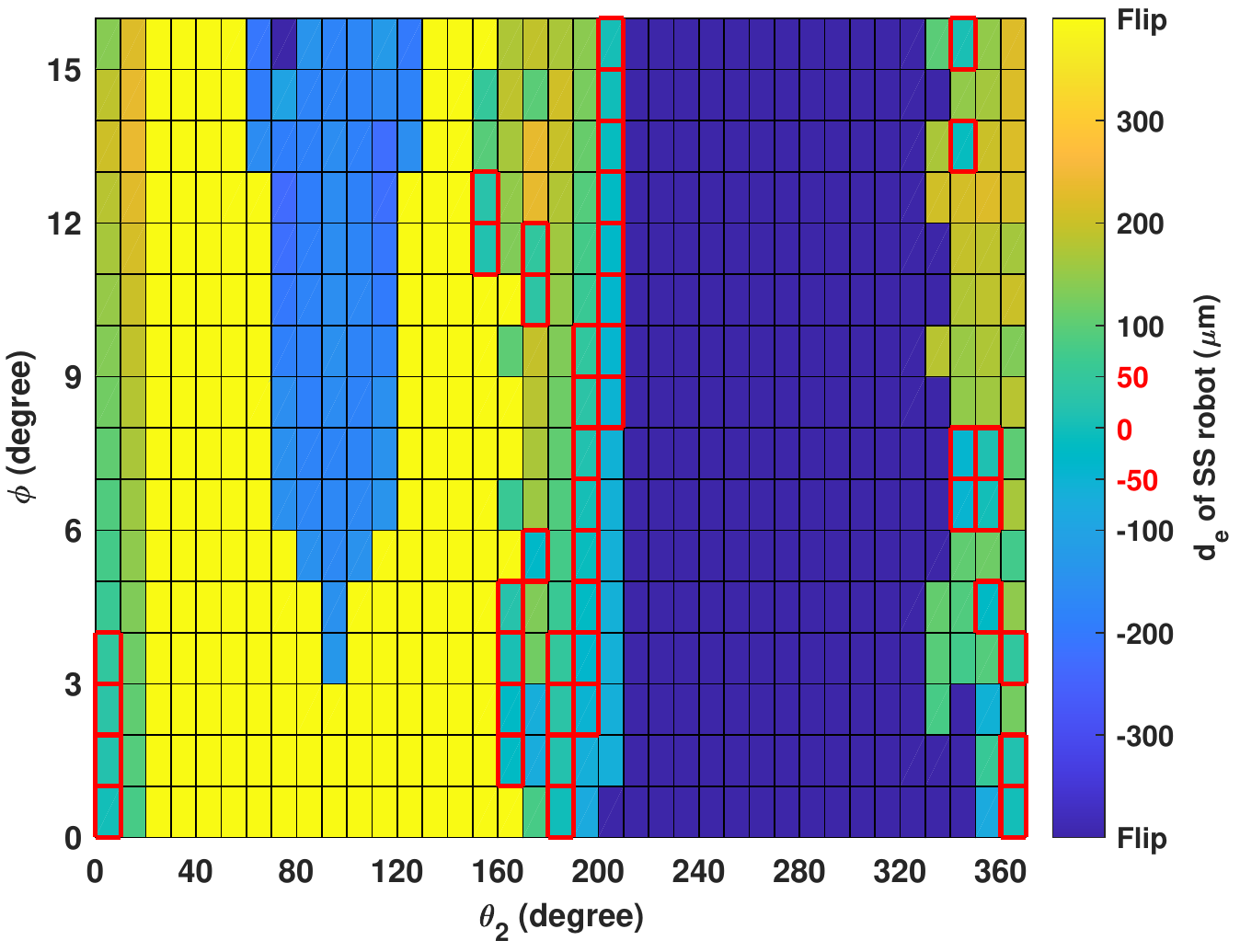}%
\caption{}
\label{analysis_d_ss_1_HZ} 
\end{subfigure}\hfill%
\begin{subfigure}[b]{1\columnwidth}
\includegraphics[width=\columnwidth]{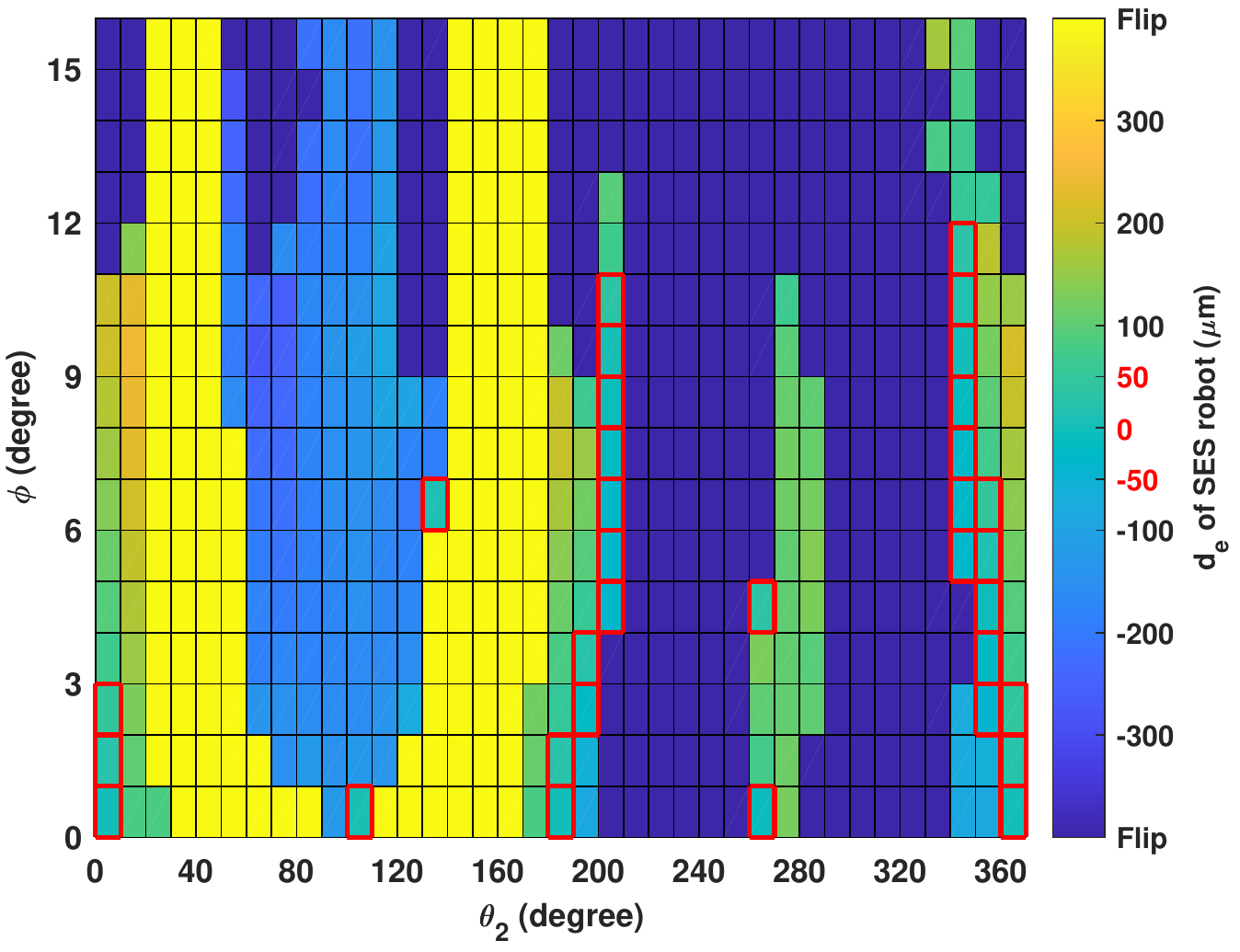}%
\caption{}
\label{analysis_d_ses_1_HZ} 
\end{subfigure}\hfill%
\begin{subfigure}[b]{1\columnwidth}
\includegraphics[width=\columnwidth]{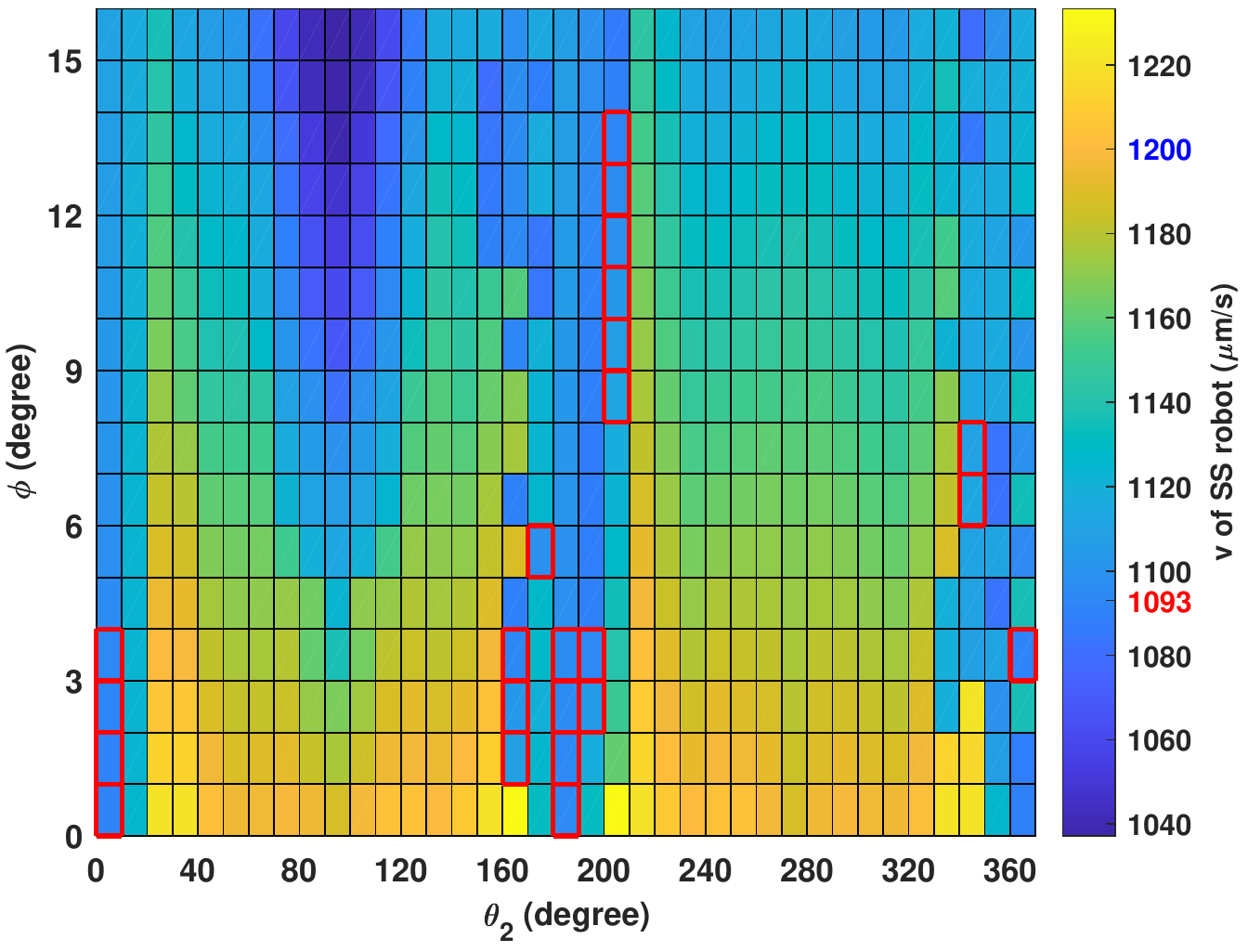}%
\caption{}
\label{analysis_ss_joint_v_1_HZ} 
\end{subfigure}\hfill%
\begin{subfigure}[b]{1\columnwidth}
\includegraphics[width=\columnwidth]{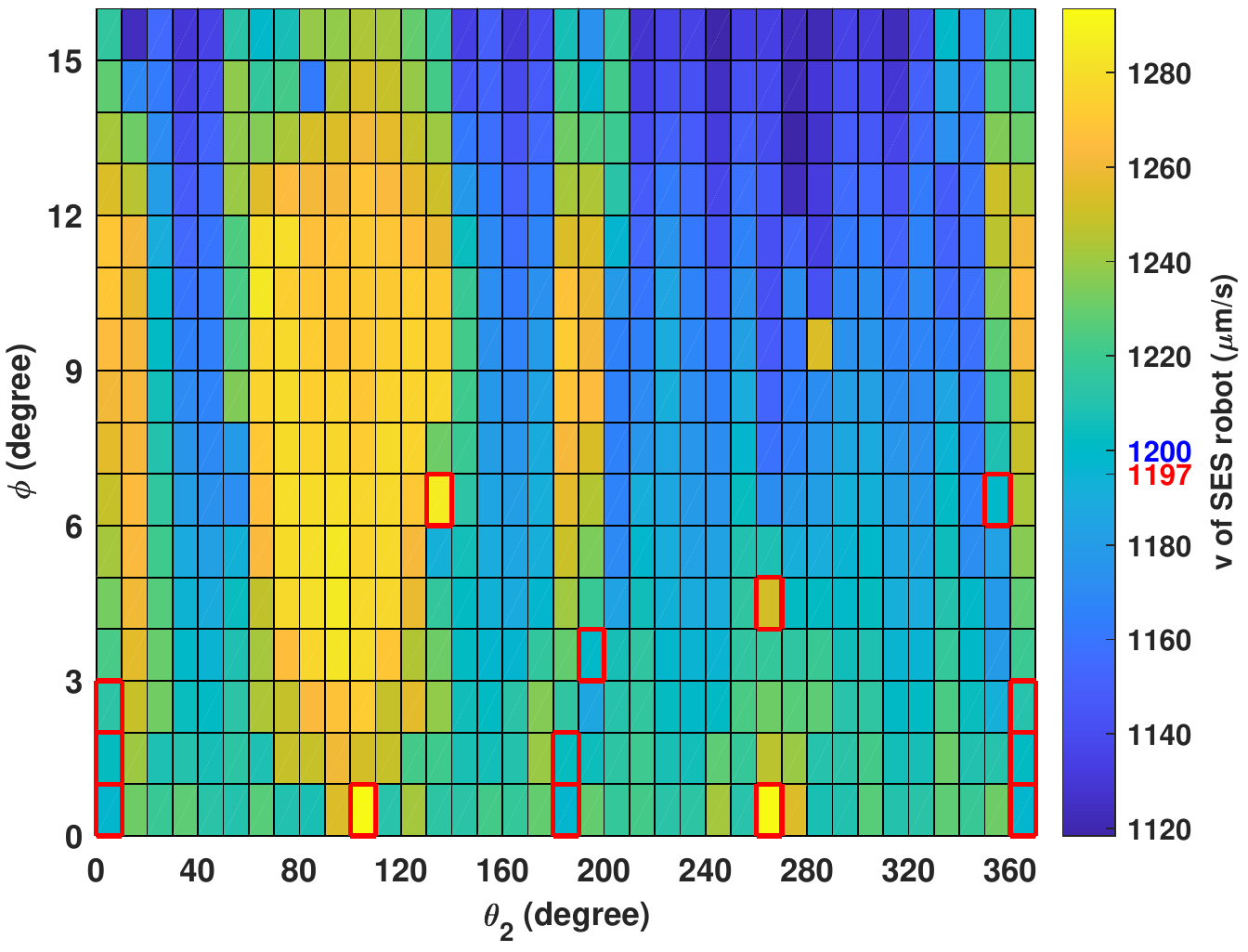}%
\caption{}
\label{analysis_ses_joint_v_1_HZ} 
\end{subfigure}\hfill
\caption{The simulation results of tumbling locomotion tests at 1 Hz with manufacturing errors $\theta_2$ and $\phi$ jointly. The plots in the first column show the distributions for SS microrobot of (a) angle of twist $\vartheta$ after one cycle of motion, (c) the drift $d_e$ after one cycle,  and  (e) the average translational speed $v$.  Similarly, the plots (b),(d) and (f) in second column show the results for SES robots. The red grids in (c) and (d) identify the cases that the robot does not flip and $|d_e|\le 50 \mu m$ after one cycle. Based on the chosen cases, the grids in (e) and (f) further identify the cases with additional condition that the speed $v$ of the robot is higher than the velocity in ideal situation. In the colorbars of (e) and (f), the value of velocity in ideal situation is shown in red, and the velocity in the situation of flipping is shown in blue. }
\label{figure:analysis_1_HZ}
\end{figure*}

\subsection{Motion Prediction}
PDMS was not observed to slip on aluminum substrates, regardless of incline angle, without large external loads. As a result, differences in incline climbing ability are minute between alternate designs of PDMS-fabricated tumbling microrobots. Therefore, we analyze the effects of the manufacturing errors and limitations by simulating their trajectory and speed on a flat surface instead. As shown in Figure~\ref{figure_magnet}, we  characterize the magnetization error with a double cone:
\begin{equation}
    f_{cone}({\bf u}) = {\bf u}\cdot{\bf u}_d -cos(\theta)\le 0
\end{equation}
where the vector ${\bf u}_d = [0,1,0]^T $ is the ideal magnetic axis, which coincides with the $\bf y$-axis of the robot. The aperture of the cone is $\theta$, and it is approximated as $\theta = 10^{\circ}$. Due to the magnetization error, the actual magnetic axis ${\bf u} \in \mathbb{R}^3$ may deviate from ${\bf u}_d$, and it should lie on or within the cone. Thus, we use two parameters $\theta_1$ and $\theta_2$ to define the unit vector ${\bf u}$:
\begin{equation*}
\begin{aligned}
{\bf u} = [\sin(\theta_2)\sin(\theta_1),\cos(\theta_1),\cos(\theta_2)\sin(\theta_1)]^T,
\end{aligned}
\end{equation*}
where $0 \le \theta_1 \le \theta $ and $0 \le \theta_2 \le 360^{\circ}$. We define the inward draft angle, $\phi$, to approximate the error in geometry of the microrobots, as shown in Fig.~\ref{figure_draft_angle}. 

\comment{
\begin{table}[!htbp]
\caption{ Parameters for PDMS-fabricated Spiked-shape (SS) and Spiked ends shape (SES) robots on aluminum.}
\begin{tabular}{c c c}
\hline \hline
Description   & Value & Units \\
\hline
Mass (m)  & $6.94\times 10^{-8}$  & kg \\
Electrostatic Force ($F_{elect}$) & $0$  & N\\
Friction Coefficient ($\mu$) & $0.54$  & -\\
Magnetic Alignment Offset ($\phi$) & $0$  & degree\\
Magnetic Volume  ($V_m$) & $3.2\times 10^{-11}$  & $m^3$\\
Magnetization ($|{\bf E}|$) & $51835$  & $A/m$\\
Coefficient of adhesion force ($C$) & 26.18 &$N/m^2$ \\
\hline
\end{tabular}
\label{table_4}
\end{table}
}

To analyze the effects of the manufacturing errors, we perform the tumbling locomotion tests in the simulation for the spiked-shape (SS) and spiked ends shape (SES) microrobots. 
The frequency of the magnetic field is chosen at $1$ Hz, and the simulation time is $1$s. In the analysis, we use the metrics as (i) the angle of twist $\vartheta$ about the longitudinal axis of the robot after one cycle of motion; (ii) The drift $d_e$ after one cycle of motion in the orthogonal direction to the direction of motion; and (iii) the average translational speed $v$ along the desired direction of motion. 

As shown in Figure~\ref{SES_tumbling}, the microrobots with manufacturing errors may twist without flipping or twist and flip to land on a side face during the tumbling motion. When the robot twists without flipping, we will say that the {\em robot twists} and when the robot twists and flips, we will say that the {\em robot flips}. We use the angle of twist after one cycle of motion, $\vartheta$, to determine whether the robot twists or flips. When the robot twists (shown in Figure~\ref{SES_tilt}), the angle $-45^{\circ} \le \vartheta \le 45^{\circ}$. When the robot flips (shown in Figure~\ref{SES_flip}), the angle $\vartheta \ge 45^{\circ}$ or $\vartheta \le -45^{\circ}$. Additionally, there exists the drift, $d_e$, in the orthogonal direction to the direction of motion.
Furthermore, when the robot twists or flips, the average tumbling speed, $v$, deviates from the ideal situation (without manufacturing errors). In the ideal situation  (shown in Figure~\ref{SES_tumb}), the microrobot lands on the protruding spikes during the tumbling motion. Based on the dimensions of the robots, the speed $v$ in the ideal situation for SS robots is $1093 \mu m/s$ and for SES robots it is $1197 \mu m/s$.  When the microrobot flips (shown in Figure~\ref{SES_flip}), it contacts with the substrate on its side planes. The speed $v$ of SS or SES depends on the length ($L = 400 \mu m$) and width ($W = 200 \mu m$) of this face. Based on Equation~\ref{model_non_slip}, the speed $v$ will be closer to $1200 \mu m/s$. 

{\bf Geometric error only}: We first analyse the effect due to inward draft angle $\phi$. The simulation results are shown in Figure~\ref{analysis_phi_1_HZ}. In the simulations, we choose $\phi = 0^{\circ},1^{\circ},...,15^{\circ}$, and the magnetic misalignment is assumed to be zero, i.e., ($\theta_1 = 0^{\circ}, \theta_2 = 0^{\circ}$). For each robot design, there are $16$ simulation runs in total.  
Let's first analyse the results of SS microrobots. The trend of the angle of twist $\vartheta$ after one cycle of motion is shown in Figure~\ref{SS_phi_pitch}.  In general, $\vartheta$ ranges from $[-20^{\circ},10^{\circ}]$, which means that during motion the robot always twists (without flipping). 
In the plot, initially the angle $\vartheta$  increases as $\phi$ increases, and it reaches the maximum value of $\vartheta = 10^{\circ}$ when $\phi = 13^{\circ}$. When $\phi =14^{\circ}$, the angle becomes $\vartheta = 0^{\circ}$, which suggests that the robot does not twist during the motion. However, the robot does twist when it elevates on the spike, and
it strikes on the flat surface after one cycle with twist angle $\vartheta = 0^{\circ}$. When $\phi = 15^{\circ}$, the robot strikes on the flat surface with $\vartheta = -21^{\circ}$ (in the opposite direction). The plot in Figure~\ref{SS_phi_d} shows the trend for the drift $d_e$ after one cycle. Note that when $\phi \ge 13^{\circ}$, the draft $d_e$ starts decreasing. The reason could be when the angle $\vartheta$ is negative, the robot drifts in the opposite direction along the orthogonal axis. This causes the overall $d_e$ to reduce. The trend for the speed $v$ along the desired direction of motion is shown in Figure~\ref{SS_phi_v}.  The speed $v$ of SS increases as the draft angle $\phi$ increases. We can conclude that the increase in draft angle $\phi$ will cause the SS robot to twist, which causes the speed $v$ and drift $d_e$ to increase. Additionally, when $\phi \le 2^{\circ}$, the drift can be relatively small: $|d_e|\le 50 \mu m$. 

We then explore the result of SES robot. In Figure~\ref{SES_phi_pitch}, the angle $\vartheta$ increases as $\phi$ increases up to $9^{\circ}$. Similar to the SS robot, when $\phi \ge  10^{\circ}$, the angle $\vartheta$ of the SES robot starts reducing until it drops to $-90^{\circ}$, indicating that the robot flips. In Figure~\ref{SES_phi_d}, the drift $d_e$ drops to zero when $\phi \ge 12^{\circ}$. In Figure~\ref{SES_phi_v}, the speed $v$  decreases when $\phi \ge 12^{\circ}$. We can conclude that the increase in draft angle $\phi$ will cause the SES robot to twist and eventually to flip. When the the flip happens, the speed $v$ starts to decrease as $\phi$ increases and the drift $d_e$ will start over
from zero. For SES robots, the draft angle needs to be $\phi \le 1^{\circ}$ in order to achieve $|d_e|\le 50 \mu m$.

{\bf Magnetization error only:} We now explore the effects only due to the magnetization error. As shown in Figure~\ref{figure_magnet}, the error cone is characterized by $\theta_1$ and $\theta_2$. Here, we restrict the possible alignments to lie on the boundary of the cone, i.e., $\theta_1 = \theta = 10^{\circ}$. (When $\theta_1 = 5^{\circ}$, the possible alignments lie within the error cone, which reduce the magnetization error in general. Due to lack of space, the results are not presented here.) There are 37 simulations runs for each robot, which includes the simulation with correct parameters ($\theta_1 = 0^{\circ}$ and $\theta_2 = 0^{\circ}$) and simulations with magnetization error ($\theta_1 = 10^{\circ}$ and $\theta_2 = 10n^{\circ}, n=1,2,...,36$).  We can observe that the plots for $v$ in Figure~\ref{analysis_theta_1_HZ} are symmetric. This symmetry is due to the fact that the geometry of the robot is symmetric. As $\theta_2$ increases, the magnetic alignment vector $\bf u$ rotates about the y-axis of the robot producing the symmetric nature of the plots.

 Again, we start with the results for SS robots. Figure~\ref{SS_theta_pitch} show the trends for the angle of twist $\vartheta$ (y-axis) vs. $\theta_2$ (x-axis). 
 When $20^{\circ} \le \theta_2 \le 160^{\circ}$ or $200^{\circ} \le \theta_2 \le 340^{\circ}$, $\vartheta$ is almost $90^{\circ}$, i.e., the robot flips. Figure~\ref{SS_phi_d} illustrates the trends of the drift $d_e$ along with the $\theta_2$. It suggests that when $\theta_2$ is near $50^{\circ}$, $140^{\circ}$, $220^{\circ}$ or $310^{\circ}$, the drift $|d_e|$ can be smaller than $50 \mu m$. However, all these cases are not preferred since the robot will flip in these instances. A detailed explanation of this is provided in Section~\ref{section_discussion}. Figure~\ref{SS_theta_v} shows the trend for speed $v$, and it illustrates that when the robot flips, the speed $v$ jumps to nearly $1200 \mu m/s$.  We can conclude that the SS robots can tumble forward without flipping, when $-10^{\circ} \le \theta_2 \le 10^{\circ}$ or $170^{\circ} \le \theta_2 \le 190^{\circ}$ (the red dots in Figure~\ref{SS_theta_pitch}).

In the case of the SES robots, Figure~\ref{SES_theta_pitch} shows that $\vartheta$ is almost zero when $-20^{\circ} \le \theta_2 \le 20^{\circ}$, $90^{\circ} \le \theta_2 \le 100^{\circ}$, $170^{\circ} \le \theta_2 \le 190^{\circ}$, and $260^{\circ} \le \theta_2 \le 270^{\circ}$ (red dots in Figure~\ref{SES_theta_pitch}). In contrast, $\vartheta$ is almost $90^{\circ}$ when $\vartheta$ is outside of these ranges. The trends of drift $d_e$ and  speed $v$ are shown in Figures~\ref{SES_theta_d}
 and~\ref{SES_theta_v} separately. When $\theta_2 = 100^{\circ}$ or $\theta_2 = 260^{\circ}$, the velocity $v$ increases to $1290 \mu m/s$ while the drift $d_e$ continues to remain close to zero. Thus, the dynamic model suggests that the overall performance of the microrobots can potentially be improved by 
 the presence of a draft angle.


{\bf Magnetization and geometric error:} We now consider the joint effect of the magnetization error and imperfections in the geometry.  Each robot shape is simulated in $592$ different runs consisting of $37$ magnetization profiles and $16$ draft angles. The simulation results are shown in Figure~\ref{figure:analysis_1_HZ}. In the plots,  the x-axis is $\theta_2$ and y-axis is $\phi$. Each grid denotes the result of the simulation run with given $\theta_2$ and $\phi$, and the color in each grid indicates the magnitude of metrics: the angle of twist $\vartheta$, the drift $d_e$, or the speed $v$.

For the SS robot, Fig.~\ref{analysis_pitch_ss_1_HZ} shows the distribution for the angle of twist $\vartheta$. When the robot flips, the grid color is yellow ($\vartheta \approx 90^{\circ}$) or dark blue ($\vartheta \approx -90^{\circ}$). Based on the plot, we can find the situations that the robot tumbles without flipping. Figure~\ref{analysis_d_ss_1_HZ} illustrates the distribution of the drift $d_e$. Similarly, we use yellow and dark blue to identify the grids in the plot where the robot flips. The red edges of the cells are used to identify the cases where the robot moves without flipping and with $|d_e|\le 50 \mu m$ after one cycle. In accordance with intuition, we see that the manufacturing errors in general cause the SS robots to drift or even flip. When both geometric and magnetization errors are small, the SS robots tend to move as desired. In some instances when geometric ($\phi$) or magnetization errors ($\theta_2$) are not small (e.g., the situation when $\phi = 15^{\circ}$ and $\theta_2 = 200^{\circ}$), the SS robots can still move as desired. Figure~\ref{analysis_ss_joint_v_1_HZ} illustrates the distribution of speed $v$ for the SS robot. We find that the speed $v$ can have higher values when the SS robot flips. This is expected since the speed for a flipped SS robot ($1200 \mu m/s$) is much higher than the ideal speed ($1093 \mu m/s$). We use red edges to further identify the cases with extra condition when the speed $v$ is higher than the velocity in ideal situation ($1093 \mu m/s$ for the SS robot). These cases can be identified when both the manufacturing errors are small. Additionally, we notice that sometimes these cases are also identified when manufacturing errors are large (e.g., $\theta_2 = 200^{\circ}$ and $8^{\circ} \le \phi \le 14^{\circ}$).
 
For the SES robot, the distributions of $\vartheta$ is shown in Figure~\ref{analysis_pitch_ses_1_HZ}. Analogous to the plot for SS robots, the SES robots tend to flip when the manufacturing errors are high. In Figure~\ref{analysis_d_ses_1_HZ}, the preferred cases for SES are also identified by the red edges. This suggests that both the SS and SES robots have similar performance. The distributions of speed $v$ for SES robots is shown in Figure~\ref{analysis_ses_joint_v_1_HZ}. We notice that the speed $v$ is often higher when the SES robot twists without flipping. It is reasonable since when the robot flips, it's speed ($1200 \mu m/s$) is almost the same as the ideal speed for SES robot ($1197 \mu m/s$). When the robot twists without flipping, the speed $v$ can be higher than the ideal speed. In Figure~\ref{analysis_ses_joint_v_1_HZ}, we  use red edges to identify the cases that satisfy the conditions that the SES robot does not flip, the drift $|d_e|\le 50 \mu m$, and the velocity $v$ is higher than the ideal speed as $1197 \mu m/s$. Among the identified cases, we can find advantageous and non-intuitive design combinations.

\begin{figure*}[!htbp]%
\centering
\begin{subfigure}[b]{0.66\columnwidth}
\includegraphics[width=\columnwidth]{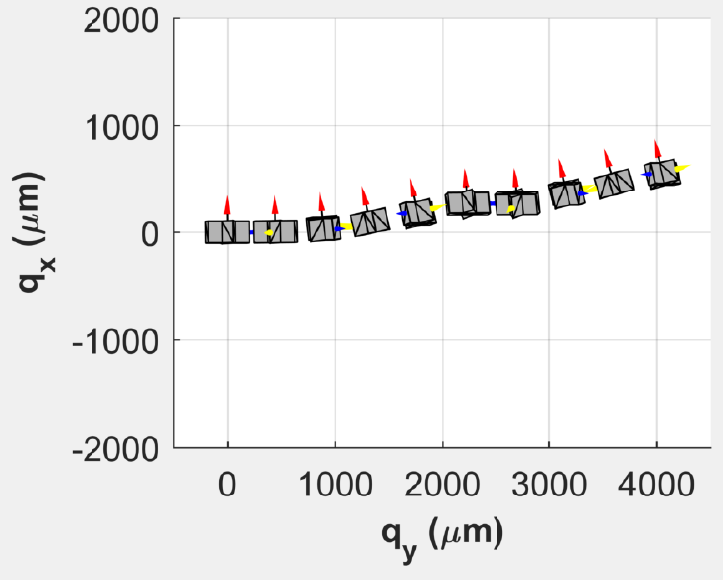}%
\caption{}
\label{snap_s_spiked_1Hz} 
\end{subfigure}\hfill%
\begin{subfigure}[b]{0.66\columnwidth}
\includegraphics[width=\columnwidth]{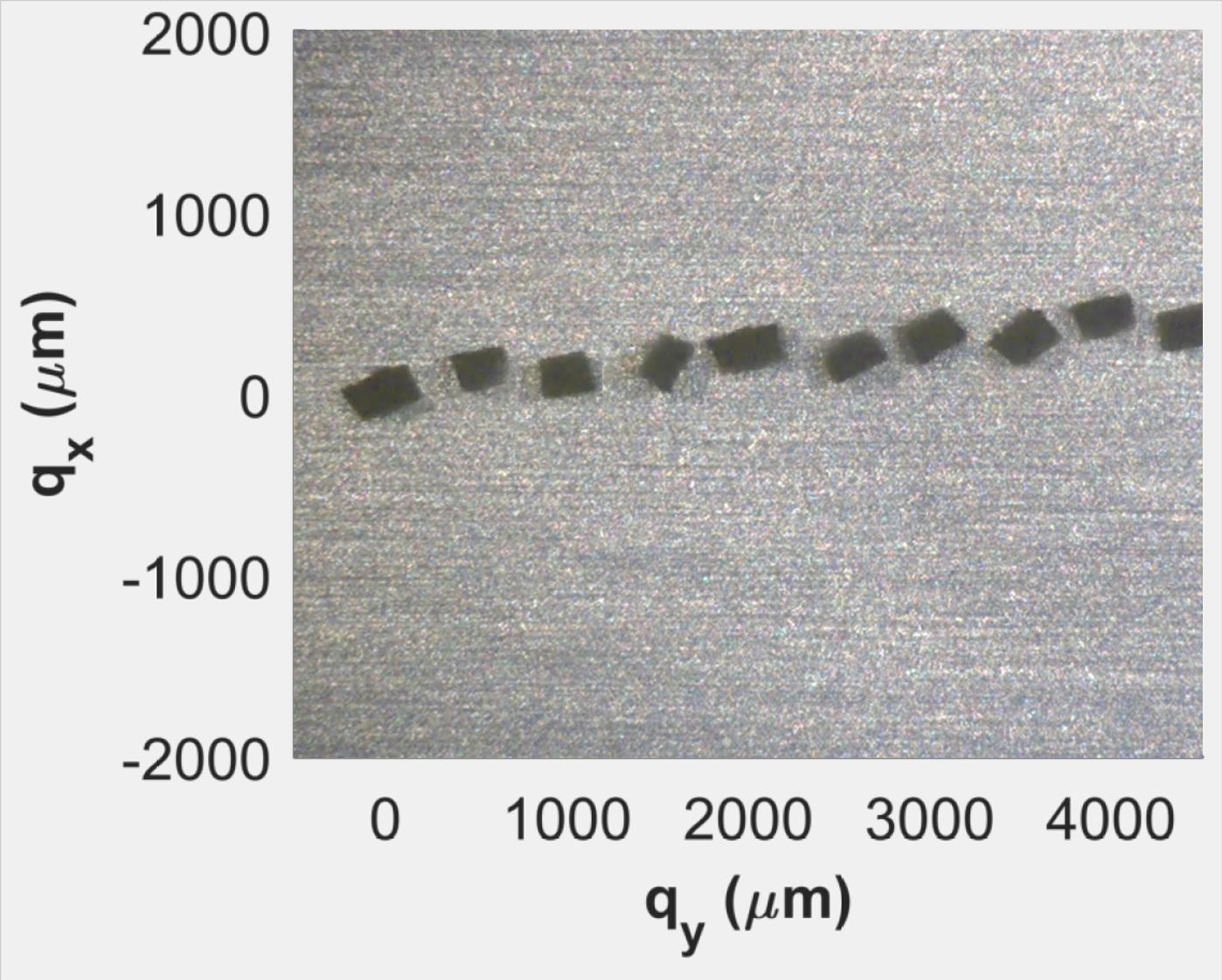}%
\caption{}
\label{snap_e_spiked_1Hz} 
\end{subfigure}\hfill%
\begin{subfigure}[b]{0.66\columnwidth}
\includegraphics[width=\columnwidth]{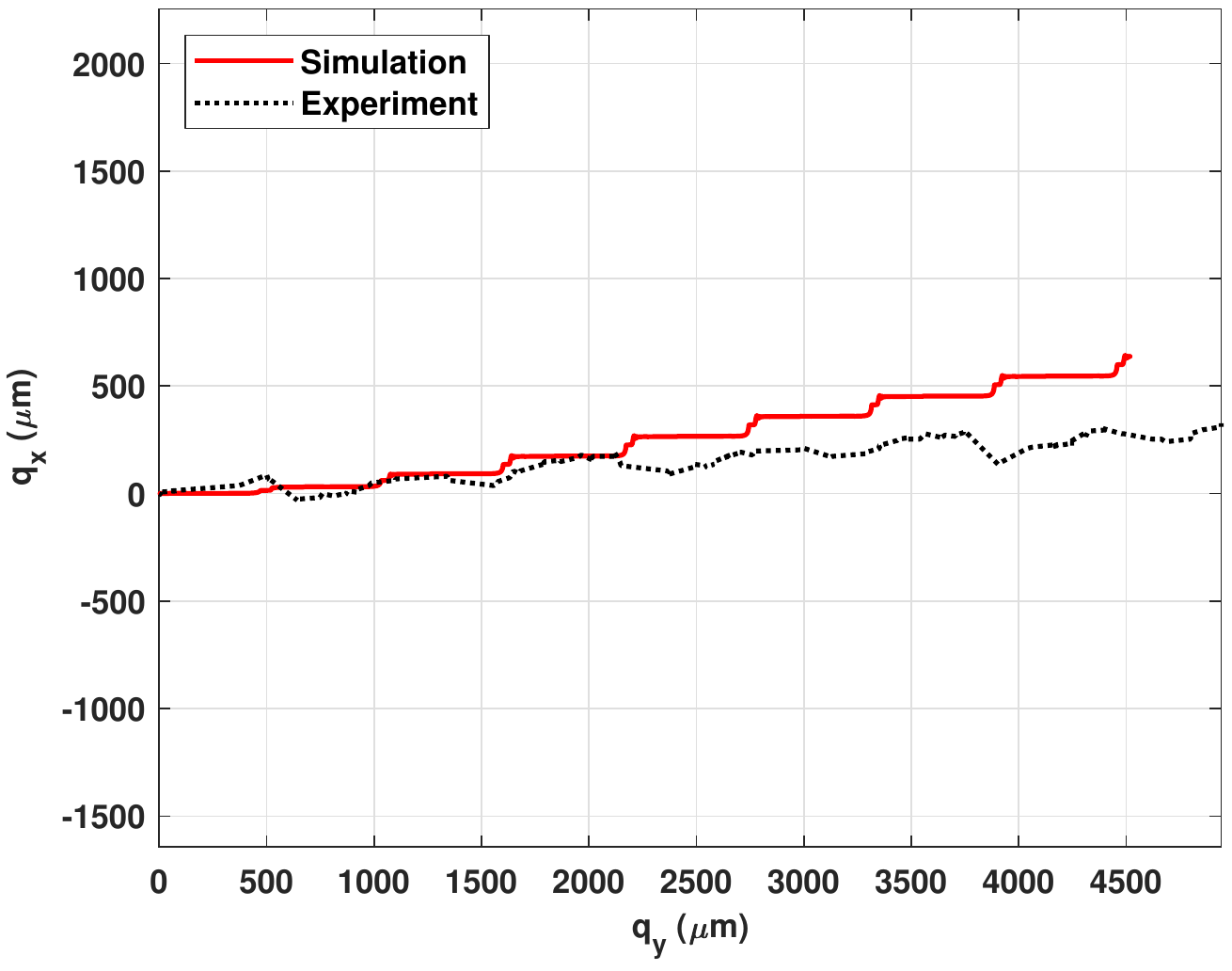}%
\caption{}
\label{trajectory_spiked_1Hz} 
\end{subfigure}\hfill%
\begin{subfigure}[b]{0.66\columnwidth}
\includegraphics[width=\columnwidth]{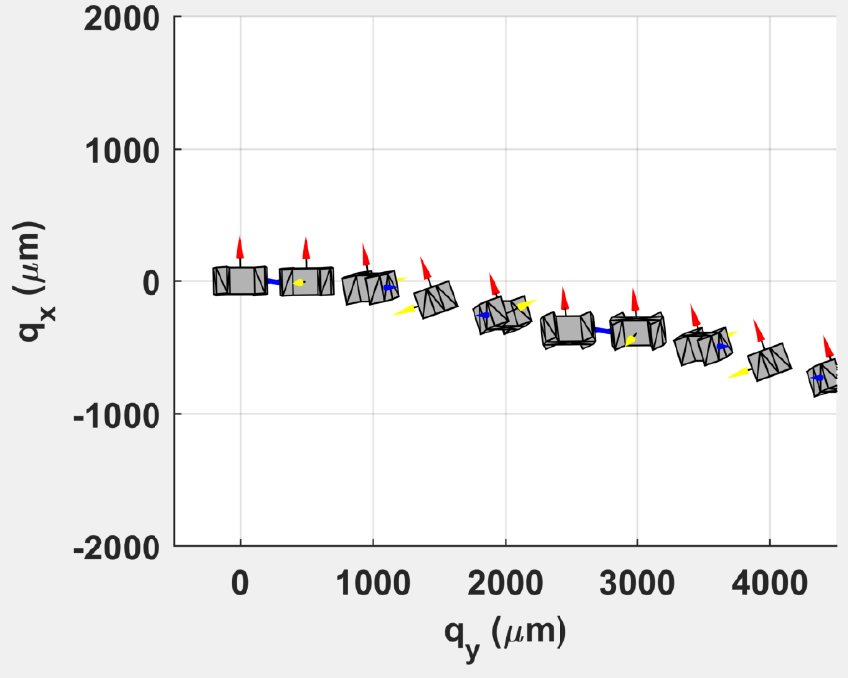}%
\caption{}
\label{snap_s_spiked_se_1Hz} 
\end{subfigure}\hfill%
\begin{subfigure}[b]{0.66\columnwidth}
\includegraphics[width=\columnwidth]{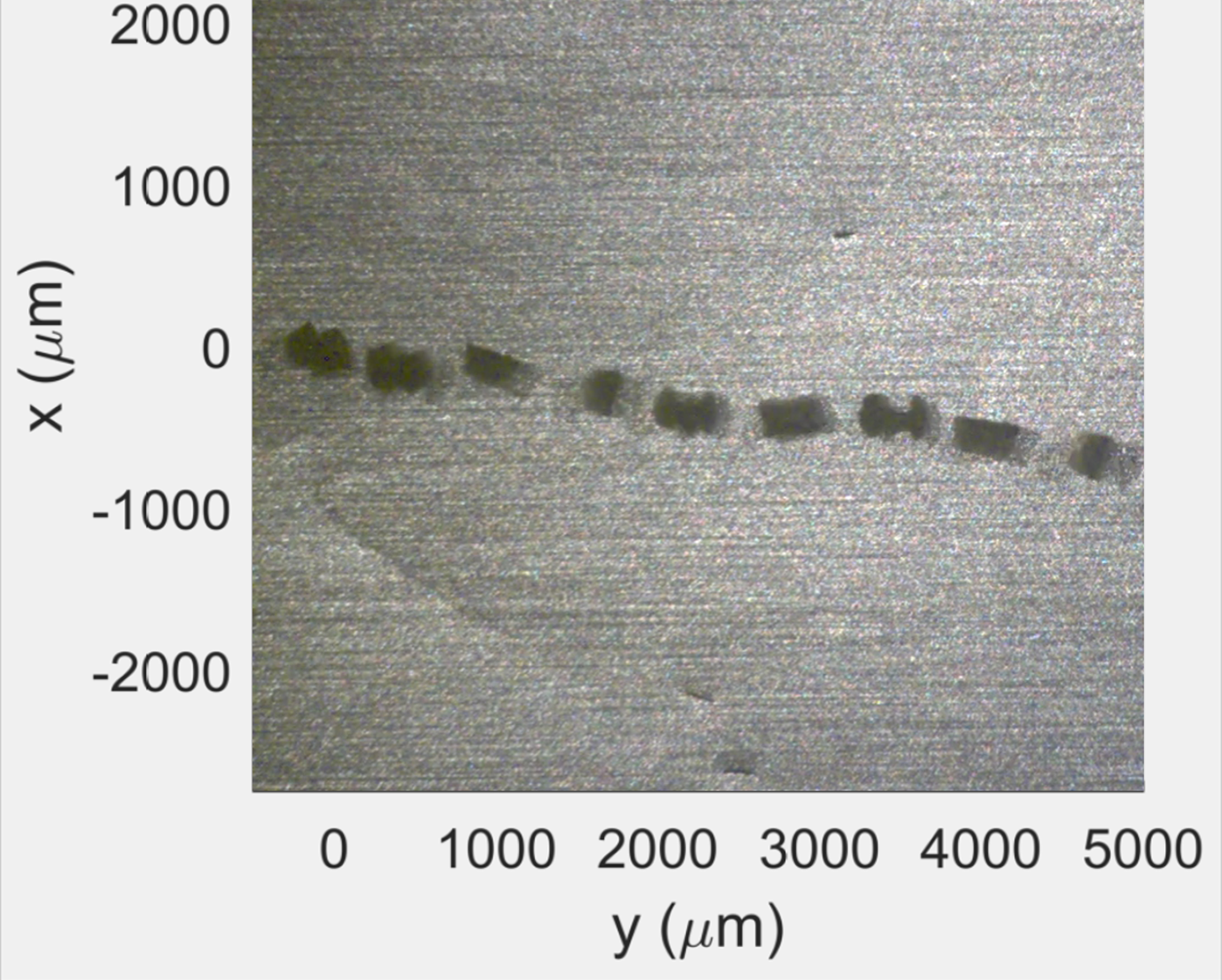}%
\caption{}
\label{snap_e_spiked_end_1Hz} 
\end{subfigure}\hfill%
\begin{subfigure}[b]{0.66\columnwidth}
\includegraphics[width=\columnwidth]{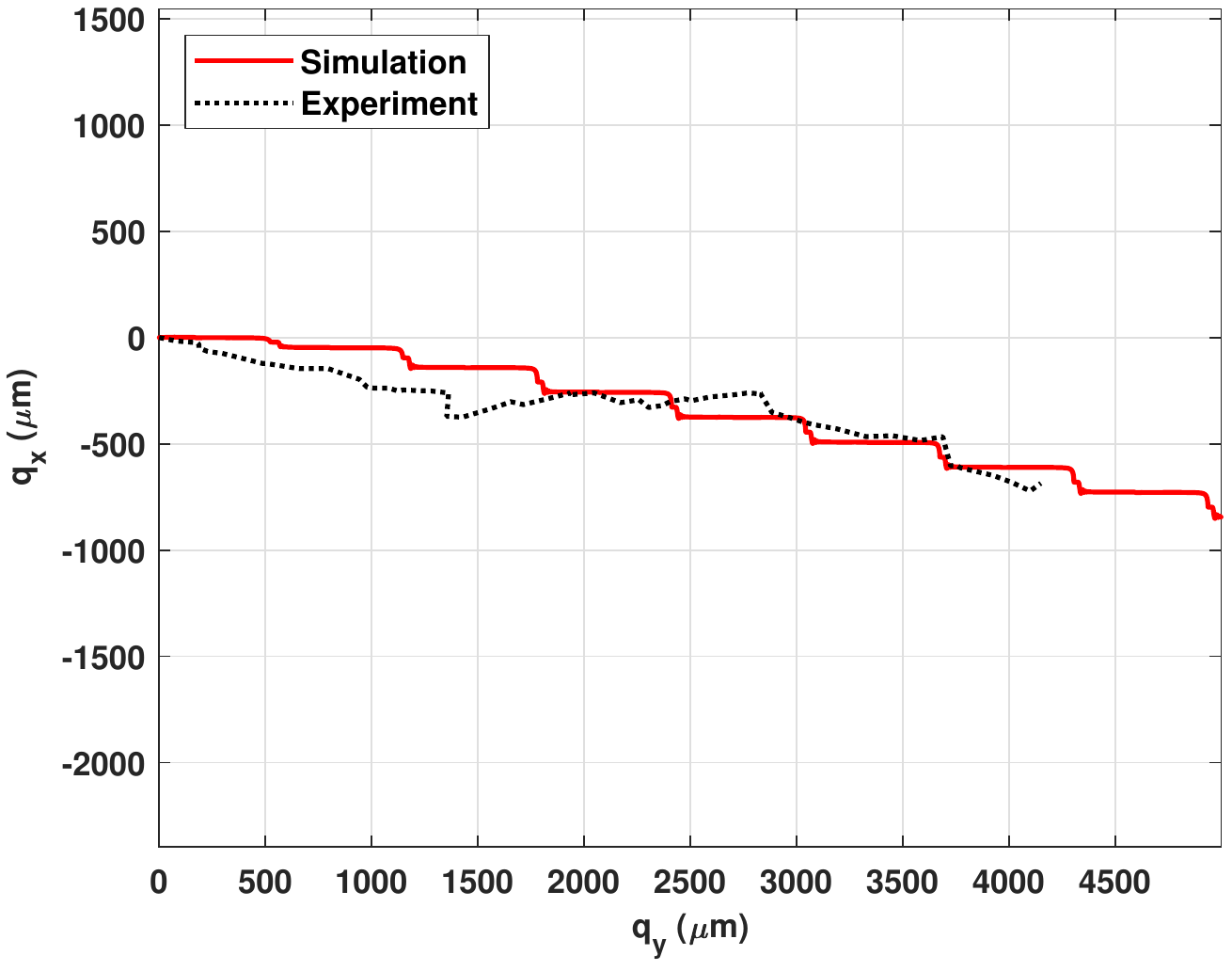}%
\caption{}
\label{x_y_spiked_end_1Hz} 
\end{subfigure}\hfill%
\caption{Trajectories of microrobots at $1$ Hz for one trial of locomotion tests on aluminium. The time for the simulation and experiment is $4$s. The first row shows the results for a SS microrobot, which includes: top view from the (a) simulation and (b) experimental results, and (c) comparison between the simulated and experimental trajectories for the X vs Y position of the microrobot. Similarly, the plots (d), (e) and (f) show the results for a SES microrobot.}
\label{figure:M_errors_1_HZ}
\end{figure*}

\subsection{Experimental Validation}

After implementing the manufacturing errors into the dynamic model, we validate the simulation with experimental results. In both simulations and experiments, we execute the locomotion tests on the substrate of aluminium.  The manufacturing error of both the SS and SES microrobots used were $\theta_1 = 10^{\circ}$, $\theta_2 = 0^{\circ}$, and  $\phi = 4^{\circ}$. The frequency of the magnetic field is at 1 Hz. The results are shown in Figure~\ref{figure:M_errors_1_HZ}. In the ideal situation, the microrobot should rotate about the rotating axis of magnetic field (without loss of generality, we choose x-axis) and tumble straight forward along the y-axis. When the manufacturing errors are taken into account, the snapshots of simulation in Figure~\ref{snap_s_spiked_1Hz} and~\ref{snap_s_spiked_se_1Hz} shows that the microrobot also drifts along the x-axis and the trajectory does not follow a straight line. Furthermore, we notice that the robot starts to twist or even flip during the tumbling cycle, changing orientation along more than one rotational axis. We then run the experiment for the same locomotion test and the results in Figures~\ref{snap_e_spiked_1Hz} and~\ref{snap_e_spiked_end_1Hz} support what we have observed in the simulation. To explore and analyze further, we plot the x-y position of CM for the SS and SES microrobots in Figures~\ref{trajectory_spiked_1Hz} and~\ref{x_y_spiked_end_1Hz} separately. 
We can conclude that the trajectory in the experiment changes periodically and the simulation matches with the experimental results with a similar pattern.

\comment{
To see the performance of the microrobots at a higher frequency, we run the simulations and experiments for both SS and SES microrobots at 2 Hz. 
To ensure the simulation and experiment processes are consistent, where the microrobots used do not change between trials, we maintain the same manufacturing error values from the 1 Hz case in the 2 Hz case. The plots for the resulting trajectories are shown in Figure~\ref{figure:M_errors_2_HZ}. In general, the simulation predicts the similar periodical shift that appears in the experiment. However, compared with the results at 1 Hz, there are larger discrepancies between the simulation and the experimental results in the 2 Hz case. The reason is likely due to the uncertainties and nonuniform surface properties of the real-world environment. While efforts were made to keep the workspace clean, small dusts particles and inherent surface roughness may impose additional loads on the microrobot that alter trajectory, amplified by positions where the microrobot is balancing about a single corner. However, due to lack of data on small scale surface variations and for the convenience of computation, the simulation model treats all substrates as a uniformly flat plane.
}
A video compilation of simulation and experimental results can be found here: \url{https://www.youtube.com/watch?v=NmxqMtOjyCg}.

\section{Discussion}
\label{section_discussion}

The intent of the alternative geometries analyzed in the previous two sections was to improve on the original cuboid geometry of the tumbling magnetic microrobots. By incorporating spiked protrusions or curved surfaces, the large faces of the microrobot would be elevated from the substrate and area contact would be minimized, reducing the effect of resistive adhesive forces. Spiked geometry was further investigated after determining this design variant would lead to the best compromise between climbing ability and translational speed. Limitations in fabrication methods, however, led to unexpected behavior that brought additional benefits and drawbacks to the more complex geometry. In order to prevent the microrobots from fragmenting during the fabrication process and during experimental handling, the constituent material was changed from doped SU-8 to doped PDMS.  Laser cutting was also used instead of photolithography to allow thicker polymeric sheets to be processed. Feature size scale and magnetic particle concentration were both reduced to making laser cutting feasible. These changes resulted in critical differences in the motion of the new microrobots and their interaction with the substrate.

Strong adhesive forces between PDMS and aluminum in dry air allow PDMS tumbling microrobots to climb inclines much steeper than the 45$^\circ$ maximum inclination angle of comparable SU-8 counterparts. PDMS is also less brittle and fragile than SU-8, making PDMS microrobots robust against large applied loads and capable of including multiple stress concentration points without breakage. When manufacturing errors are minimized, spiked geometry microrobots do not encounter area contact during the tumbling cycle, reducing the minimum magnetic torques necessary to actuate the microrobots. In practice, fabrication through laser cutting introduces tapered edges that result in an inherent draft angle on the sides of the PDMS microrobots. Based on Figure~\ref{analysis_phi_1_HZ}, larger draft angles generally result in increased translational velocities at the cost of a proportionally scaling drift away from the intended straight-line motion. Through the dynamic model, it is predicted that this drift can be kept at 50 $\mu m$ or less when the draft angle $\phi$ is $\le 2^\circ$ or $\le 1^\circ$ for the SS robots and SES robots, respectively.

Manufacturing errors in the magnetization of the microrobots can occur regardless of their constituent material or geometry. These errors are introduced from the manual alignment and mounting of the microrobots during the magnetization process. SS and SES PDMS microrobots are more susceptible to misalignment than cuboid SU-8 variants due to their multiple protrusions and compressibility, making mounting difficult. The resulting misalignment leads to magnetic torques that cause rotation/twisting along unintended directions. This problem is further compounded when draft angles are included, where point contact is frequent and the microrobot has less resistance against spinning or flipping to the side. When considered in combination, it is estimated from the dynamic model that the absolute magnetization error should be kept at $|\theta_2| \le 10^\circ$ and the absolute draft angle at $|\phi|$ $\le 2^\circ$ in order to ensure the resulting microrobot drift is 50 $\mu m$ or below.

Upon inspection of the simulation results in Figure~\ref{analysis_phi_1_HZ} and Figure~\ref{analysis_theta_1_HZ}, there appear to be several advantageous parameter variations where velocity increases without a proportional increase in drift. In Figure~\ref{analysis_theta_1_HZ}, for example, the SS robot's translational velocity increases to $1210 \mu m/s$ when the magnetization error $\theta_2$ is 150$^\circ$ while the drift continues to remain close to zero. This behavior suggests that intentionally introducing manufacturing errors can potentially lead to better overall performance for the microrobot. It is important to point out, however, that the magnitude of the twist angle $\vartheta$ is greater than $45^{\circ}$ in the majority of these disproportionate cases and the microrobot tumbles in the `flip' orientation. This `flipped' tumbling orientation has a larger outer perimeter along the side profile, resulting in higher translational speeds, but the microrobot also experiences significantly more area contact in this orientation. Instead of balancing over the spiked protrusions, as intended, the microrobot is periodically striking the substrate with the flat surfaces of its side planes. Due to the high adhesive forces between PDMS and aluminum, this frequent area contact may result in the microrobot getting stuck against the substrate, with the actuating torque not strong enough to counteract the increased adhesion. This effect is observed to occur in practice and should be considered in tandem with potential improvements in speed and drift. From Figure~\ref{analysis_theta_1_HZ}, we also note that in one particular case, when the magnetization error $\theta_2$ is 100$^\circ$ or 260$^\circ$ for the SES robot, the speed increases to $1290 \mu m/s$ with minimal drift and twisting introduced. Thus, through comprehensive variation of simulation parameters, advantageous but non-intuitive design combinations can be found. A caveat is that this occurs when magnetization error is considered in the absence of geometric error. In practicality, there is often a mixture of non-zero magnetization and geometric errors.

While predictions from the simulation model do not match exactly with experimental results, the overall qualitative trends are similar between the two data sets. Without needing to spend significant time and resources on iterative physical prototyping of microrobots, new geometries and parameter variations can be rapidly analyzed to help find superior design combinations. Improved, but non-intuitive combinations can be found, as discussed earlier, where introducing certain manufacturing errors could lead to potential improvements in performance. By estimating the manufacturing tolerances necessary to keep drift below a maximum level, the dynamic model can help determine whether minimizing magnetic alignment error or draft angle error is more cost-effective. The combination of PDMS material and laser cutting fabrication comes with limitations and drawbacks, but also introduces a larger geometric design space from which to combat those faults. Our model is well-posed to reduce the resources necessary to explore this design space and make further improvements to microrobot design.
\section{Conclusions}

In this paper, we have demonstrated a dynamic simulation model that can account for intermittent non-point contact over multiple substrates and surface inclinations. We validated this model using experiments incorporating a tumbling magnetic microrobot and predicted that spiked ends geometry would result in better overall performance. Using the model as a design aid would help save time and reduce costs on the microrobot iteration and fabrication process. Despite manufacturing errors and limitations in the fabrication of more complex geometries, we show that the simulation model successfully can reproduce the effects of these errors for further predictions. Future developments may include accommodations for soft, elastomeric robot bodies without necessitating a rigid body assumption and additional modeling for wet environments.
\section{Acknowledgements}

The authors would like to acknowledge Seunghwan Jo and Martin Byung-Guk Jun for their assistance with laser cutting and access to their lab resources. The authors would also like to acknowledge the facility access at Birck Nanotechnology Center (Purdue University) and Georges Adam for his assistance with microrobot fabrication.

\bibliographystyle{asmems4}


\appendix       

\end{document}